\pdfoutput=1

\documentclass[sigconf]{acmart}

\AtBeginDocument{%
  \providecommand\BibTeX{{%
    \normalfont B\kern-0.5em{\scshape i\kern-0.25em b}\kern-0.8em\TeX}}}

\usepackage{caption,subcaption}
\usepackage{multirow}

\begin{document}

\title{(Un)fairness in Post-operative Complication Prediction Models}




\author{Sandhya Tripathi}
\authornote{Department of Anesthesiology, Washington
              University in St Louis, MO, USA}

\author{Bradley A. Fritz}
\authornotemark[1]

\author{Mohamed Abdelhack}
\authornotemark[1]

\author{Michael S. Avidan}
\authornotemark[1]

\author{Yixin Chen}
\authornote{Department of Computer Science and
              Engineering, Washington University in St Louis, St Louis,
              MO, USA}

\author{Christopher R. King}
\authornotemark[1]

\begin{abstract}
With the current ongoing debate about fairness, explainability and transparency of machine learning models, their application in high-impact clinical decision-making systems must be scrutinized. 
We consider a real-life example of risk estimation before surgery and investigate the potential for bias or unfairness of a variety of algorithms. 
Our approach creates transparent documentation of potential bias so that the users can apply the model carefully. 
We augment a model-card like analysis using propensity scores with a decision-tree based guide for clinicians that would identify predictable shortcomings of the model. 
In addition to functioning as a guide for users, we propose that it can guide the algorithm development and informatics team to focus on data sources and structures that can address these shortcomings. 
\end{abstract}



\maketitle

\section{Introduction}
Biases and irrationality in human decision making based on irrelevant or socially unacceptable factors has been observed in essentially every context \cite{levinson_forgotten_2006} with substantial variation based on personal background \cite{hamberg_gender_2008}. 
Although the reasons and mechanisms are different, decision-making algorithms have shown similar biases.
Potential explanations include bias encoded in the choice of training data, choice of data collection, the impact of miss-classified outcomes and missing input data, the effect of group size, and the chosen objectives \cite{gianfrancesco2018potential}.
An instructive example on the power of these biases came in 2016, \cite{angwin2016machinePropublica} when a detailed analysis of COMPAS, a recidivism prediction algorithm that has been used in United States to aid pre-trial holding decisions, found that it was disadvantaging black defendants. 
Based on these observations, fairness and transparency is a critical evaluation to make when deploying machine learning for clinical decision making \cite{vollmer2020machineQuestions}. 
For example, if the system made unacceptable predictions selectively for racial minorities, its application could increase health disparities and even worsen overall outcomes compared to no system at all \cite{rajkomar_ensuring_2018,goodman2018machine,fairness_prec_med,obermeyer2019dissecting,nordling2019fairer,vyas2020hiddenNEJM}.
Suppose there is no true difference in outcome between the majority and minority groups. Consider a system that accurately identifies high or low risk in the majority group, but never flags members of the minority as high risk. Implementation of such a system would shift healthcare resources from the minority group to the majority group, potentially increasing disparities.
Empirical studies have demonstrated these issues in actual application: gender imbalance leading to biased classifiers \cite{larrazabal2020gender}, in general medicine and mental health care \cite{chen2019canAI}, in renal failure prediction models \cite{williams2019towards}, classifying chest x-rays \cite{seyyed2020chexclusion}, and in clinical contextual word embeddings \cite{zhang2020hurtful}.

One such system for clinical decision assistance without explicit decision support was described by \cite{fritz_deep-learning_2019,cui_factored_2019}, which developed machine-learning classifiers for postoperative mortality and other complications based on preoperative risk factors and real-time intra-operative data.
The explicit goal of this program was directing attention from telemedicine providers \cite{king_protocol_2019} to the ``neediest'' patients, with similar systems used to stratify patients into risk-appropriate surgery during preoperative clinic visits and to inform them of potential adverse events \cite{bilimoria_development_2013}.
A major obstacle to using this system is the need to understand if it makes ``fair predictions,'' or if it has ``bias'' with respect to protected or sensitive  characteristics. 
For clarity of prose, a sensitive or protected characteristic refers to the membership information of societal groups like race or sex, that are protected by anti-discrimination law \cite{barocas2016big,civil_rights1964}. 

There are three major conceptual variations of fairness which have been proposed for use with predictive algorithms: disparate treatment, disparate impact and disparate mistreatment.\cite{zafar2019fairness} 
An algorithm suffers from:
\begin{itemize}
    \item \textbf{Disparate treatment \cite{barocas2016big}} if it provides different outputs for groups of people with the same (or similar) values of non-sensitive features but different values of sensitive features. This is also referred as direct discrimination \cite{pedreshi2008discrimination}.  
    \item \textbf{Disparate impact \cite{barocas2016big}} if it provides outputs that benefit (or hurt) a group of people sharing a value of a sensitive feature more frequently than other groups of people \textit{marginal on the values other features take}. Also referred as statistical parity \cite{corbett2017algorithmic} or demographic parity \cite{dwork2012fairness}.
    \item \textbf{Disparate mistreatment \cite{zafar2017fairnessWWW}} if the algorithm's performance differs for otherwise similar people at different levels of a sensitive feature. Also referred as equality of opportunity \cite{hardt2016equality} and predictive equality \cite{corbett2017algorithmic}. This approach may offer equal utility in some contexts \cite{rajkomar_ensuring_2018}.
\end{itemize}

We will consider covariates as a potential explanations, and do not consider disparate impact as it requires proportionality in decision outcomes for sensitive groups that could have arisen from justifiable reasons (e.g. if women have lower rates of smoking).
For the remainder of the paper, we will refer to ``bias'' as disparate treatment and ``unfairness'' as disparate mistreatment / differential benefit.
We argue that unfairness is most relevant at the preoperative clinic stage of our application, where the most important use is to guide patients to appropriate preoperative procedures and inform them of the risks of surgery.
However, in the telemedicine stage of the application, bias also becomes relevant.
Even if a group were truly at systematically lower risk, we may not want that group to receive a strongly reduced amount of clinician attention, especially as clinicians may react in a non-linear way to the risk estimates.

\subsection*{Addressing and Preventing Bias}
A review and comparison of existing fairness interventions can be found in \cite{friedler2019comparative,parikh2019addressing}. 
However, reducing or removing bias can come at the cost of decreased predictive performance. 
For example, flexible classifiers ought optimize performance within and across groups; imposing equal accuracy may simply force the algorithm to degrade its predictions in the ``easier'' category \cite{chen2018mydisClas}. 
Examples of enforcing minimal bias while optimizing the model performance have nevertheless been attempted where context-specific justifications apply, \cite{zafar2019fairness,cho_fair_ISIT,ghassami2018fairnessInf,adel2019one}
including several clinical examples \cite{pfohl2019counterfactual, pfohl2020empirical, pfohl2019creating}.
Without contextualizing to domain specific needs, blind application of off the shelf approaches can worsen differential performance \cite{corbett2018measureCriticalreview,herington2020measuringUnjust,fazelpour2020algorithmic} and create other problems \cite{veinot2018goodintentions}. 

In contexts where eliminating bias is impossible, an alternative approach is to transparently document the problem and educate users so that they can apply the model with care. \cite{mitchell2019modelcards} 
Our approach is closely related to \cite{mitchell2019modelcards}; 
after initial empirical demonstration of unfairness on ML models in a real example in Section \ref{sec: evid_for_bias}, in Section \ref{sec: passive_app} we extend the tabular presentation of accuracy in different groups with a decision-tree based guide which identifies for clinicians where the model has predictable shortcomings.
With this knowledge, clinicians can act accordingly on the predicted risk for a particular patient, the algorithm development team can focus innovation on data sources and structures likely to address present shortcomings, and informatics and clinical partners can improve data collection efforts among underserved populations.


\section{Dataset and prediction task} \label{sec: prob_descr}
Postoperative complication prediction models rely on the information collected during preoperative (pre-anesthesia clinic) assessments contained in the electronic health record.  
Access to the data was approved by the Human Research Protection Office at Washington University in St Louis, USA with a waiver of informed consent (IRB number $201607122$).
Data definitions and inclusion criteria were selected to match prior reports. \cite{fritz_deep-learning_2019,cui_factored_2019}
Data for patients age at least 18 years who underwent surgery with tracheal intubation at Barnes-Jewish Hospital (St Louis, MO, USA) between June 2012 and August 2016 are included in the current study. 
The preoperative assessments contain lists of past medical history items, demographics, anthropometrics, laboratory values, vital signs, integrative wellness assessments (ASA-PS values), and some lifestyle factors (smoking, heavy drinking).
For each laboratory test, the most recent value before surgery (but no more than 30 days old) is utilised.
For this example, we use only preoperative features, mimicking a tool that could be used in preoperative clinic or in day-of-surgery planning.
A more complete description of the predictors is including in \cite{fritz_deep-learning_2019,cui_factored_2019, king_association_2020}.
Continuous features were normalized to $[0,1]$ as a pre-processing step. 
Values that fell outside the plausible ranges for continuous features (provided by domain experts) were interpreted as input errors and set to missing and were later imputed using the mean value over the dataset.  
For categorical features, all values that did not exist in the data dictionary of possible values were labelled as missing and combined into a new category. 

The two prediction tasks we focus on are post operative 30 day mortality and Acute Kidney Injury (AKI). 
Acute kidney injury is defined according to KDIGO guidelines for changes in serum creatinine \cite{noauthor_acute_nodate}; an increase in the serum creatinine value by $\ge 0.3$ mg/dL or $\ge$50\% within 48 hours is considered positive.
Patients with preoperative renal failure were excluded from AKI prediction models.
Sensitive features for our example were taken to be self-reported sex and race.
Because of the very small number of individuals reporting sex as other than female or male, only these two categories are shown.
Similarly, the dataset contains a relatively small number of individuals reporting race other than black or white; we present results only for the black/white contrast.
The outcome rates and marginal distribution of sensitive features are given in Table \ref{tab: feature_dist_marginal_dist}. 

As can be seen in Table \ref{tab: feature_dist_marginal_dist}, the outcome rate is highly imbalanced for both the targets, which could decrease the performance of the prediction models. 
To reduce the effect of imbalance, we used an approach of cost sensitive learning with domain expert given costs for class specific mis-classification and down sampling \cite{elkan2001foundations}.
We used negative and positive class mis-classification cost pair value of $(1,25)$ and $(1,14)$ for 30 day mortality and AKI respectively. 


We checked for the existence of bias or unfairness across a range of classifiers, namely, Decision Trees (DT), Random Forests (RF), Gradient Boost (GB), Logistic Regression (LR) and a Deep Neural Network (DNN). All these classifiers were implemented using the \textit{sklearn} module available in Python. Most of the arguments of these classifiers were implemented with default values except for the following parameters: \{DT: max\_depth =20\}, \{RF: n\_estimators=200\}, \{GB: n\_estimators=200, max\_depth=1\}, \{DNN: hidden\_layer\_sizes=(256,256), solver='sgd', alpha=8e-5, batch\_size=64, learning\_rate\_init=0.01, max\_iter=10\}. In main text, we only report the difference between the performance measures for RF and LR classifiers. The actual performance measure values for all classifiers is provided in the figures in Supplementary Material (SM).

Performance measures examined for unfairness are False Negative Rate (FNR), False Positive Rate (FPR), False Discovery Rate (FDR) and False Omission Rate (FOR).
We present bootstrap estimates and standard errors of performance metrics using 150 bootstrap samples using the out-of-bag error as a smoothed version of k-fold cross validation. \cite{efron1983estimating}
Because of the reduced effective sample size within each bootstrap replicate, the CV error will tend to overestimate that which would be obtained with training on the entire dataset, but this is at least consistent within each experiment.
We present the results in mean and confidence interval format rather than a hypothesis testing one.
In the case of discrimination, the magnitude of differences is critically important in addition to their statistical certainty; we would be interested in large differences even if there was remaining uncertainty, and we would be willing to de-prioritize very small differences even if highly confident that they were non-zero.

Let $\mathcal{X} \subseteq \mathbb{R}^{n}$ be the feature space and $\mathcal{Y} = \{-1,1\}$ be the label set. 
Let $\mathcal{D}$ be the joint distribution over $\mathbf{X}, \mathbf{Z}, Y$ with $\mathbf{X}, \mathbf{Z} \in \mathcal{X}$ , $Y\in \mathcal{Y}$ and $\mathbf{Z}$ be the set of sensitive features. 
$\mathbf{Z}$ generally consists of variables taking ordinal or nominal values. 
In practice, however, we are only provided with an i.i.d. sample of size $m$ from distribution $\mathcal{D}$, viz., $ D = \{(\mathbf{x}_i, \mathbf{z}_i, y_i)\}_{i = 1}^{m}$.

\begin{table*}[]
\begin{tabular}{|c|c|c|c|c|c|c|c|c|}
\hline
\textbf{} & \textbf{Outcome} & \textbf{Total} & \textbf{Female} & \textbf{Male} & \textbf{Other Sex} & \textbf{\begin{tabular}[c]{@{}c@{}}Black \end{tabular}} & \textbf{\begin{tabular}[c]{@{}c@{}}White \end{tabular}} & \textbf{Other Race} \\ \hline
\textbf{\begin{tabular}[c]{@{}c@{}}$\#$ of samples\\ (full dataset)\end{tabular}} & \textbf{Both} & 52499 & 26206 & 25897  & 396 & 10207 & 39838 & 2454 \\ \hline
\multirow{2}{*}{\textbf{Outcome rate}} & \textbf{30 day mortality} & 0.035  & 0.027 & 0.042 & 0.038 & 0.027 & 0.034 & 0.085 \\ \cline{2-9} 
 & \textbf{AKI} & 0.063  & 0.048 & 0.078 & 0.040 & 0.065 & 0.061 & 0.082 \\ \hline
\end{tabular}
\caption{First row reports the marginal distribution for subgroups of the sensitive variables. Last 2 rows report the outcome rates for subgroups of sensitive variable.} 
\label{tab: feature_dist_marginal_dist}
\end{table*}

\section{Empirical examination of bias} \label{sec: evid_for_bias}



We first document whether our classifiers exhibit bias.
As demonstrated in Table \ref{tab: feature_dist_marginal_dist}, there are notable differences between men and women in marginal rates of death and AKI, and smaller differences between white and black participants.
Participants with "Other" or missing race had substantially higher rates of death and AKI, but this is difficult to interpret because ``missing'' can encode ``declined to answer'', ``no applicable category'', or informative missingness during e.g. an emergency surgery in which the patient is unable to answer.

An important point is whether the difference in outcomes is explained away by differences in other variables or stems from direct use of the sensitive feature.
For a given patient $i$ from test set  whose data is denoted by $(\mathbf{x}_i,\mathbf{z}_i,y_i)$ and for a trained predictor $f(\cdot)$, we compare the predictions made for patient $i$ for the true value of the sensitive feature and for the complementary value of the sensitive feature, i.e., $f(\mathbf{x}_i,\mathbf{z}_i,y_i)$ and $f(\mathbf{x}_i,1-\mathbf{z}_i,y_i)$. 
We also compute the predicted values using classifiers trained not including the sensitive characteristic.
Finally, we evaluate the classifiers on a \textit{propensity score matched} sample.
That is, in order to compare only individuals with similar covariates, we calculate $e_i = P(\mathbf{Z}_i =1|\mathbf{X}_i)$ and 1-to-1 match individuals with $z=1$ to individuals with similar $e$ and $z=0$.
Because the propensity score, is a balancing score, after conditioning on it, the distribution of measured baseline covariates (features other than sensitive features) is similar between ``treated'' (sensitive feature $\mathbf{Z}_i = 1$) and ``untreated'' (sensitive feature $\mathbf{Z}_i = 0$) subjects (patients) \cite{austin2011introduction}.

We estimate the propensity score by using a logistic regression model where the sensitive feature is regressed on observed baseline features ($\mathbf{X}$), although more flexible classifiers can also be used. 
While there are many methods of using the propensity score to accomplish balance, we chose propensity score matching (PSM) as several studies demonstrate that it eliminates a greater proportion of the systematic differences in baseline characteristics between the subgroups of sensitive feature \cite{austin2011introduction}, is simple to implement, and can be applied to all classifiers easily.
With a caliper value of $0.05$ in PSM, our test set did not have any match for 2.8\% and 69\% of the disadvantaged (Z=1) samples when conditioned on race and sex respectively, reflecting major differences in distributions of procedures performed and risk factors. 
As expected, matching across sex was more difficult as there are a fair number of sex-specific surgeries (gynecologic surgeries, prostate surgery) or nearly sex-specific surgeries (breast surgery); patients undergoing these procedures essentially do not contribute to the estimates in the propensity matched sample. 
Using a relatively transparent classifier to generate propensity scores also helps to identify any features which are essentially surrogates for the sensitive characteristic (such as gynecologic surgery).
In our example, we found that ideal body weight also acted as a surrogate (because it is a deterministic function of sex and height) and was excluded from propensity score calculation; height also contributed strongly to group separation, but was retained in the model because of its potential importance in other calculations.

\begin{figure*}[h!]
    \centering
\includegraphics[width=1\textwidth]{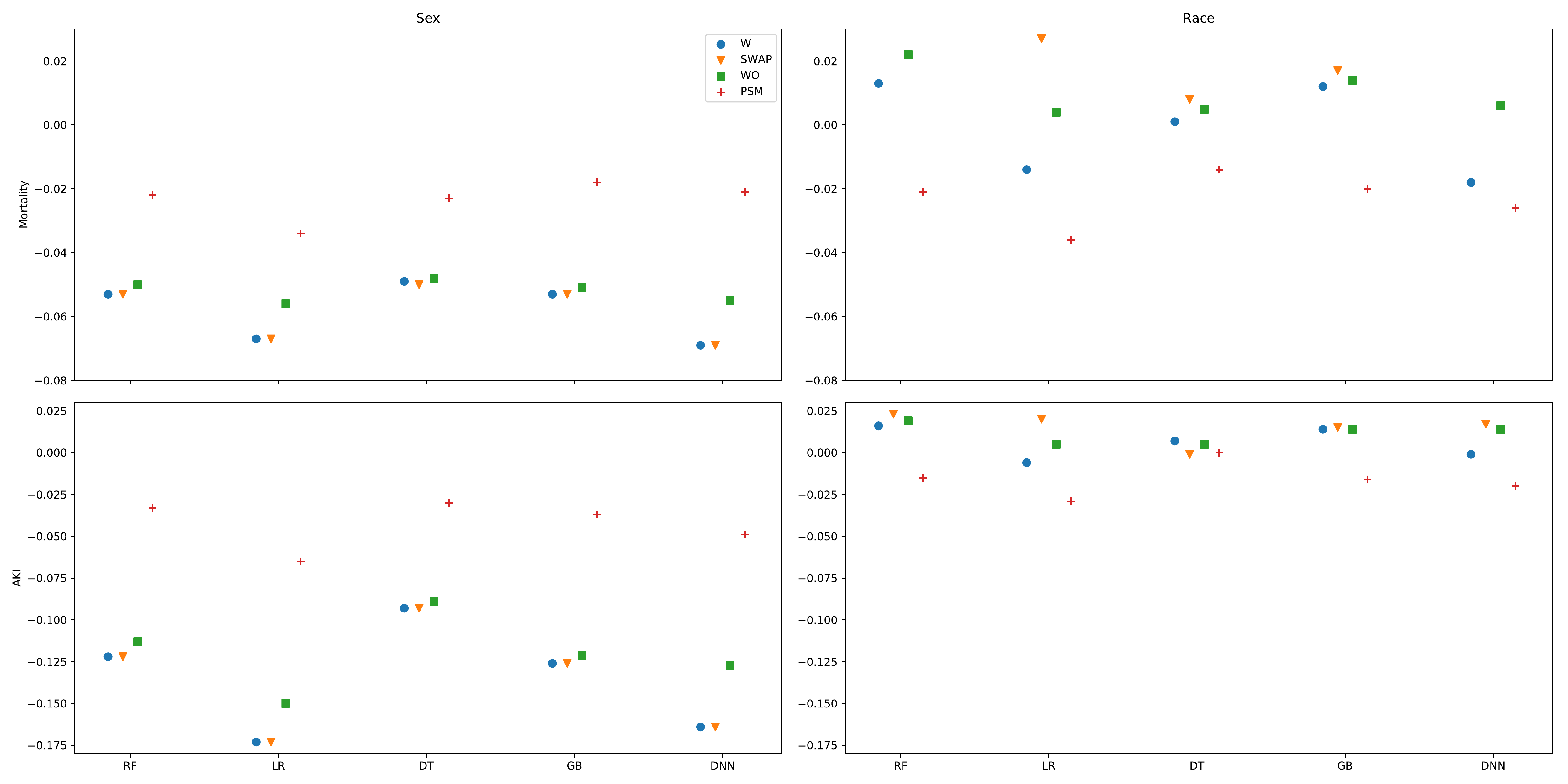}
    \caption{ Classifier predicted outcome (30 day mortality and AKI) rates by sensitive characteristic (Sex and Race). Inclusion of sensitive characteristic in learning set (W), average outputs with sensitive characteristic inverted (SWAP), sensitive characteristic omitted from learning set (WO), and restricted to propensity score matched sample (PSM). Each panel depicts the combination of an outcome and sensitive characteristic $(Z)$ with y-axis values being the difference in predicted outcome rates between the subgroups $Z=1$ and $Z=0$.
    }
    \label{fig: classifier_bias}
\end{figure*}

In Figure \ref{fig: classifier_bias} we examine the influence of the sensitive characteristic on average outputs.
We see that the differences with respect to sex and race observed in Table \ref{tab: feature_dist_marginal_dist} are maintained.
These differences persist or are even magnified when excluding sex as a feature or swapping the sex inputs.
Sex differences are greatly reduced in the propensity matched sample.
These suggest that observed characteristics (covariables) correlated with sex largely explain the differences in average predicted values.
The smaller differences in predicted outcomes across race are similarly unchanged by scrambling or omitting race, but are eliminated (or reversed in sign) in the propensity matched group, suggesting that observable covariables explain some of the differences in outputs, but that unobserved factors directly explain some as well.
Similar patterns were observed across classifiers.


\begin{figure*}[h!]
    \centering
    \begin{subfigure}[b]{0.478\textwidth}
        \centering
        \includegraphics[width=1.06\textwidth]{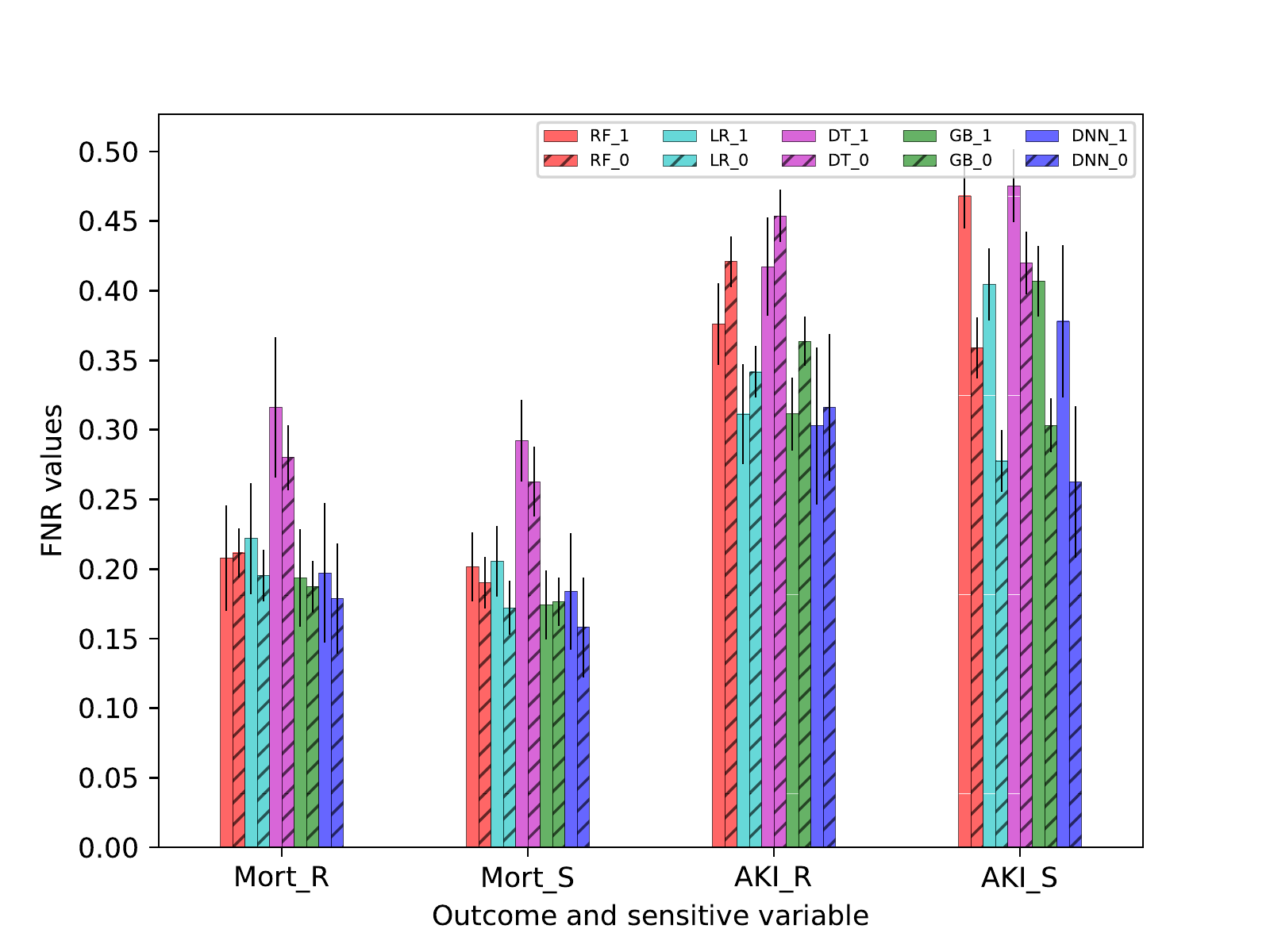}
    \end{subfigure}
    \hfill
    \begin{subfigure}[b]{0.478\textwidth}  
        \centering 
        \includegraphics[width=1.06\textwidth]{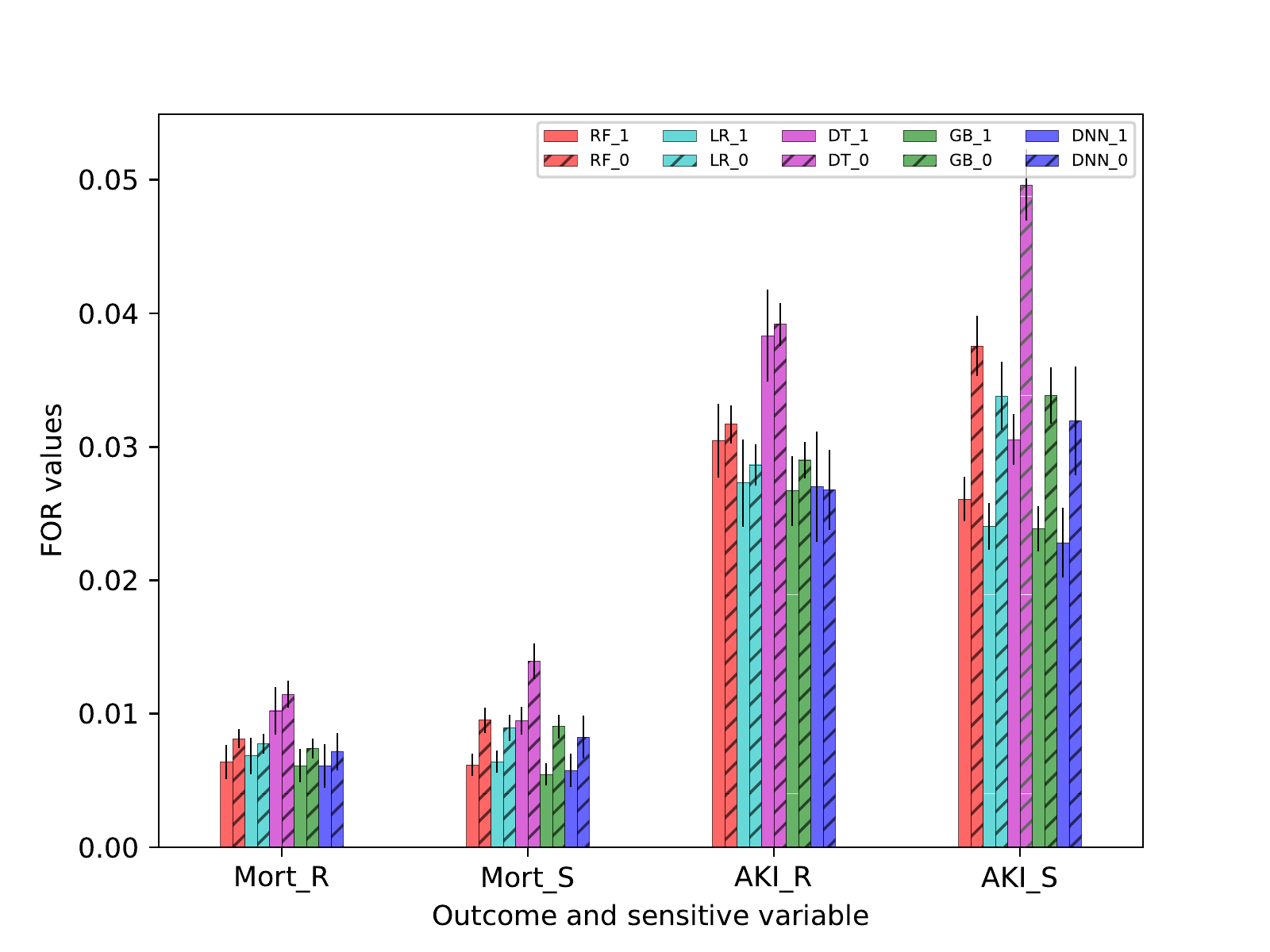}
    \end{subfigure}
    \vskip\baselineskip
\begin{subfigure}[b]{0.478\textwidth}
        \centering
        \includegraphics[width=1.06\textwidth]{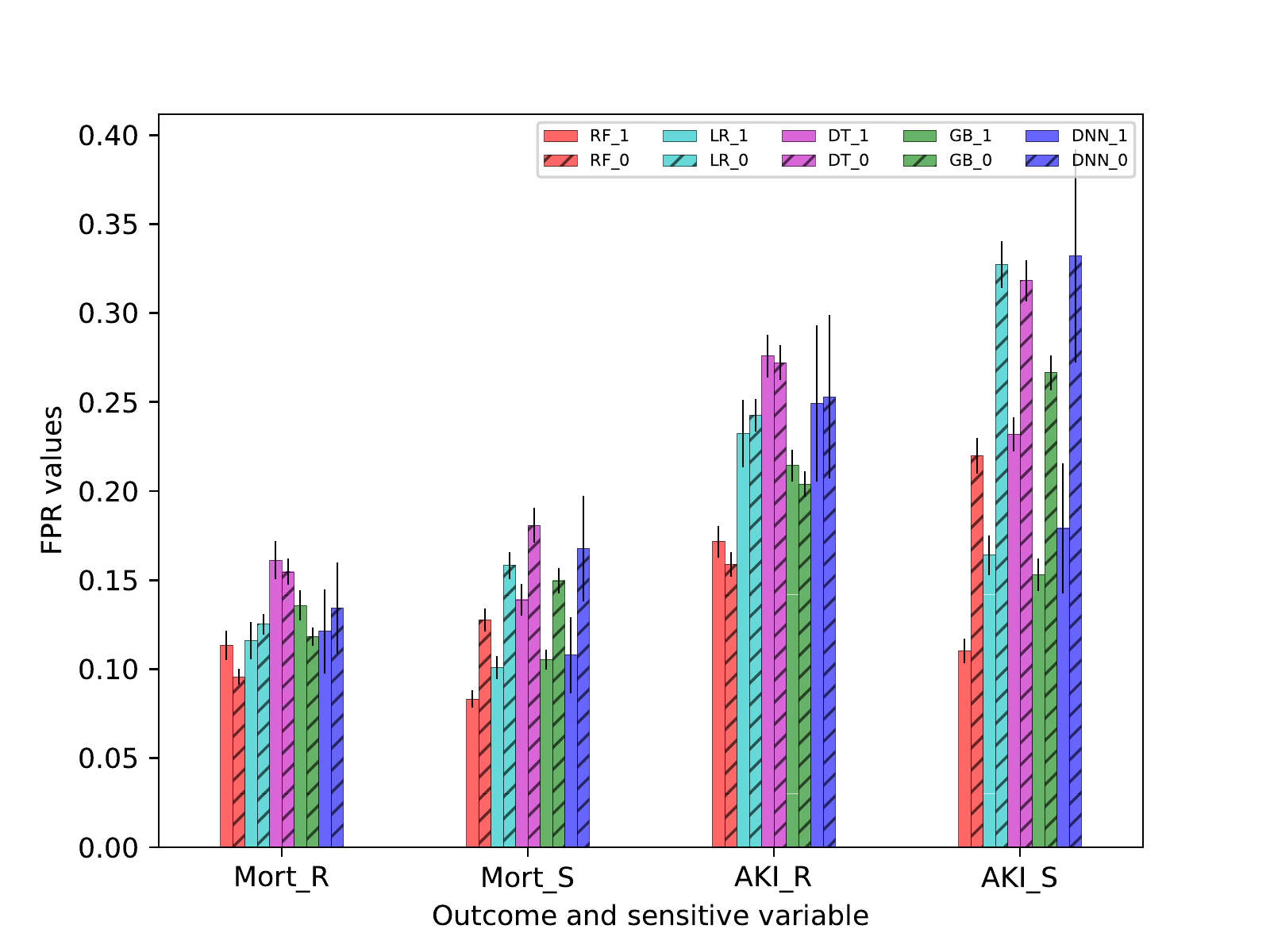}
    \end{subfigure}
    \hfill
    \begin{subfigure}[b]{0.478\textwidth}   
        \centering 
        \includegraphics[width=1.06\textwidth]{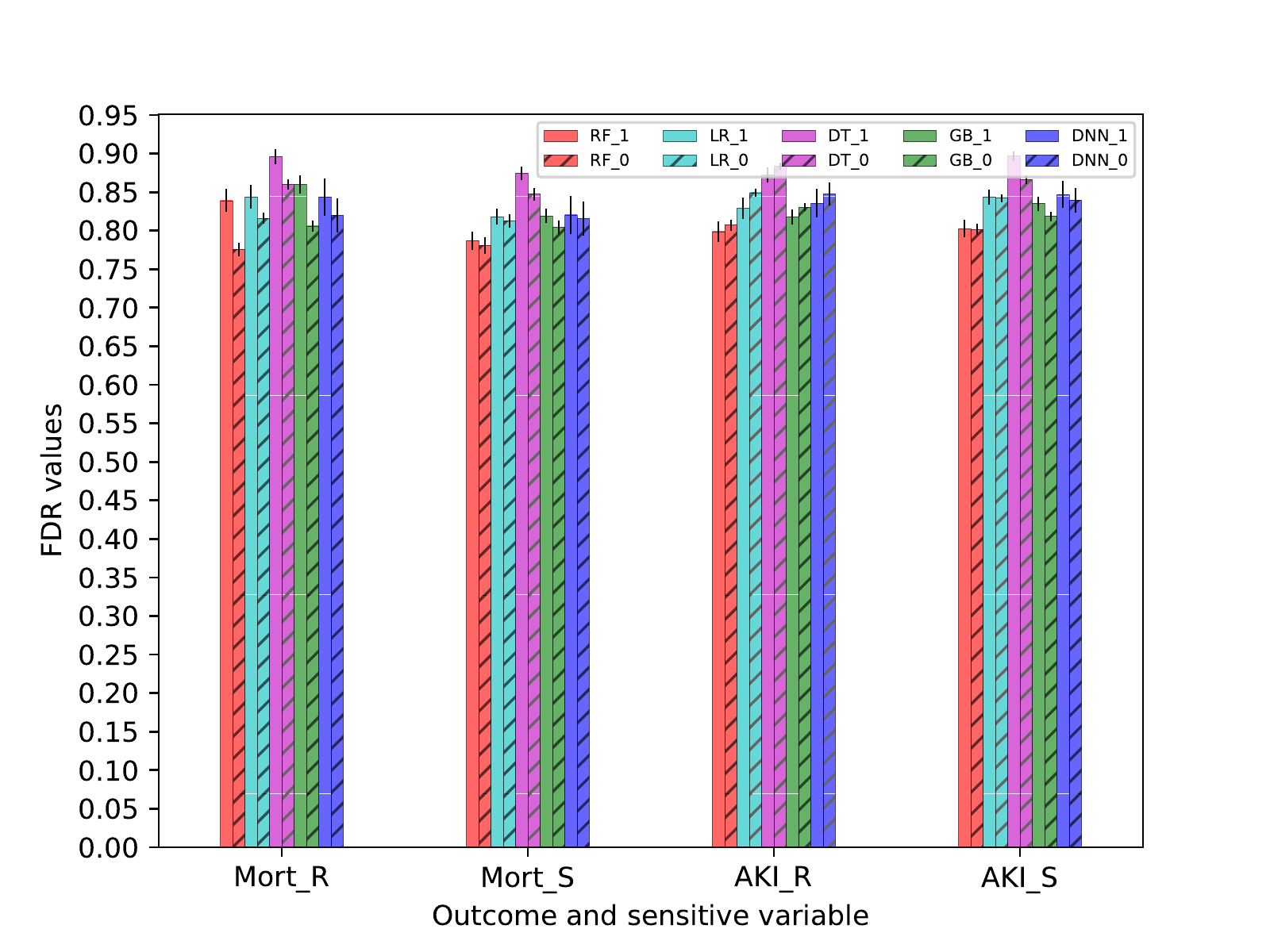}
    \end{subfigure}
    \caption{Each panel in the figure corresponds to a performance measure $\in$ {FNR, FOR, FPR, FDR} for the classifiers $\in$ {RF, LR, DT, GB, DNN} marked by color across subgroups (1 = solid bar or 0 = hash marked bar) of sensitive characteristic Race (R) and Sex (S) for outcome 30 day mortality (Mort) and AKI. For sensitive characteristic Sex, 1 denotes females and 0 denotes males; for Race, 1 denotes Black and 0 denotes White. A high value of FNR or FOR (FPR or FDR) implies discrimination against (benefits for) the population belonging to that group of protected attribute.
    }
    \label{fig: basic_models_diff_perf }
\end{figure*}

We next turn to unfairness in the form of differential predictive accuracy.
For each subgroup of a sensitive feature, we separately computed the test set performance of the models trained on sample $D$ with criterion $\in$ \{FNR, FPR, FDR, FOR\}.
These measures are shown in Figure \ref{fig: basic_models_diff_perf }.
No meaningful differences are observed in AKI performance across race.
However, all classifiers have markedly higher FNR and lower FPR among women.
While these measures are natural inputs to a utility calculation, they are affected by the baseline rate of the outcome.
FDR shows no meaningful differences across sex, but the FOR remains lower among women.
For mortality prediction, RF and GB show modestly higher FPR among black participants, and all classifiers show higher FDR among black participants.
No differences are observed across race in FNR or FOR.
Women have equivalent FNR and slightly better FOR across all classifiers, and substantially lower FPR with essentially equivalent FDR.


\subsection{Algorithmic sources of bias}
To explain these differences, we first consider purely algorithmic effects related to group size and information sharing.
Although the total sample size is reasonably large, the number of adverse events is modest, and we could easily be in a regime where smaller groups (such as black race) have not ``saturated'' the more flexible classifier's ability to learn from the provided features, leading to worse performance in minority groups.
In a closely related effect, classifiers may learn patterns in the larger majority group which do not discriminate as well in the minority groups (interaction of other features with the sensitive feature).
We explore this possibility in three ways.
First, we downsample the larger group to an equal number of observations as the smaller group.
Second, we fit classifiers in stratified samples, preventing any carry-over of pattern recognition.
Third, we compare classifiers with different degrees of freedom.
We would expect carry-over effects to be the worst in low-degree of freedom classifiers and group size effects to be the worst in high degree of freedom classifiers.

\begin{figure*}[h!]
    \centering
    \begin{subfigure}[b]{0.478\textwidth}
        \centering
        \includegraphics[width=1\textwidth]{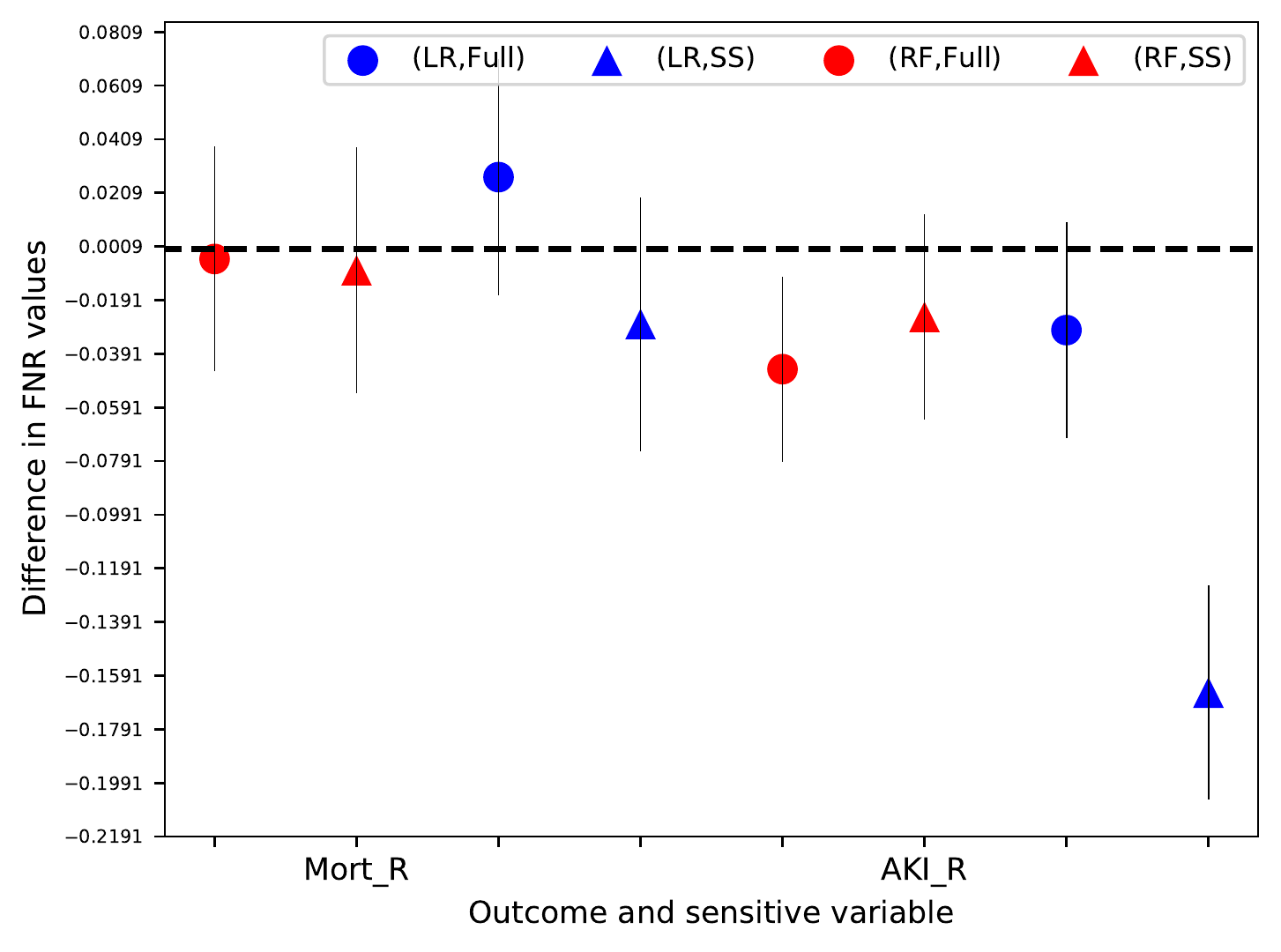}
    \end{subfigure}
    \hfill
        \begin{subfigure}[b]{0.478\textwidth}
        \centering
        \includegraphics[width=1\textwidth]{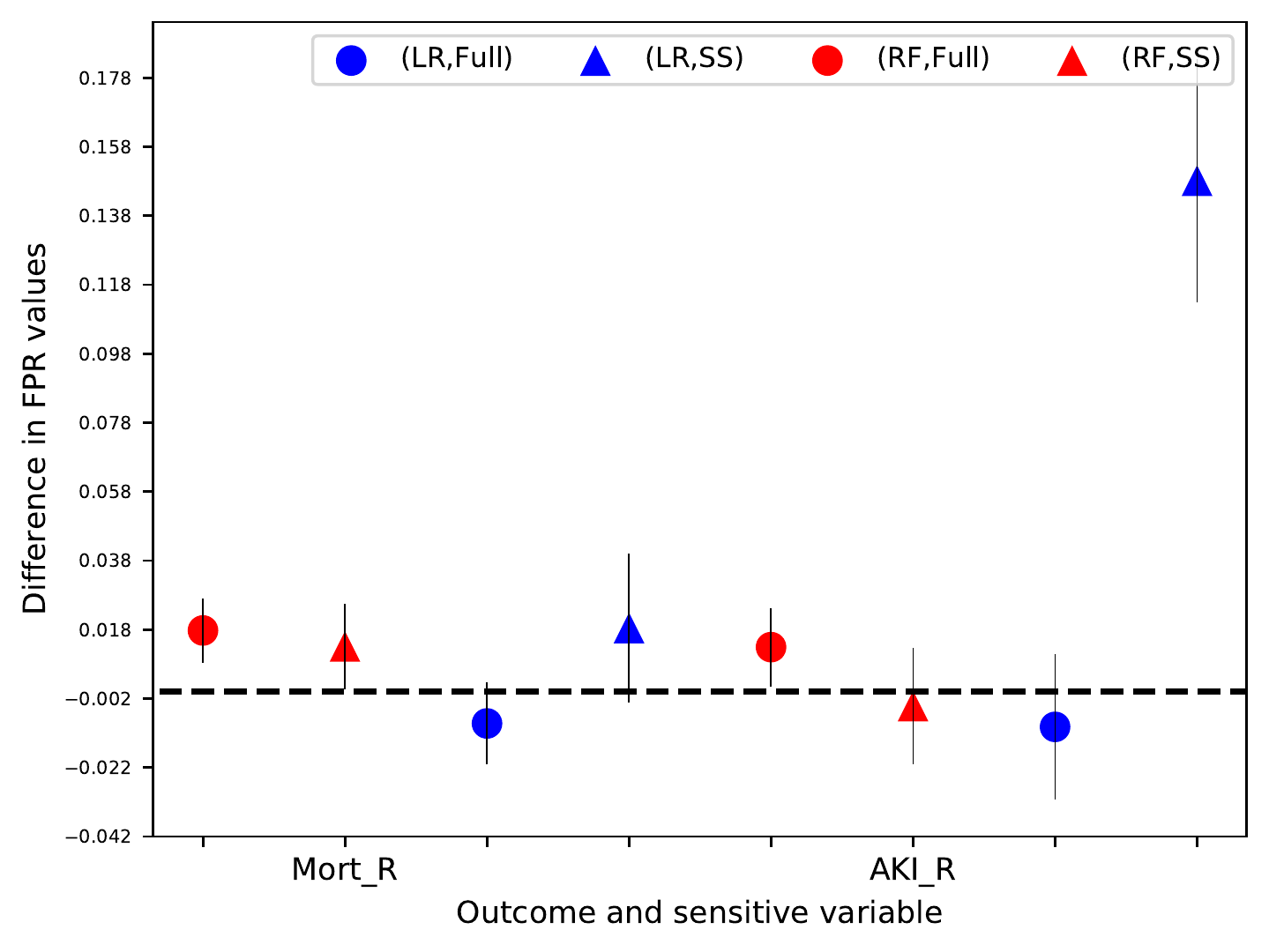}
    \end{subfigure}
    \caption{Figure comparing the difference in performance measure values $\in$ {FNR, FPR} across subgroups of sensitive characteristic Race (R) when the models were trained on down-sampled majority subgroup data for outcomes 30 day mortality (Mort) and AKI. Full and SS denotes training on full and down-sampled data. Down-sampling did reduce the FNR for people belonging to black race but increased it for white race which is not a good solution. Actual values of FNR, FPR along with FOR and FDR for all classifiers can be found in SM Figures \ref{figapp: subs_models_diff_perf_FNR_FPR} and \ref{figapp: subs_models_diff_perf_FOR_FDR}.}
    \label{fig: subs_models_diff_perf}
\end{figure*}

\begin{figure*}[h!]
    \centering
    \begin{subfigure}[b]{0.8\textwidth}
        \centering
        \includegraphics[width=0.85\textwidth]{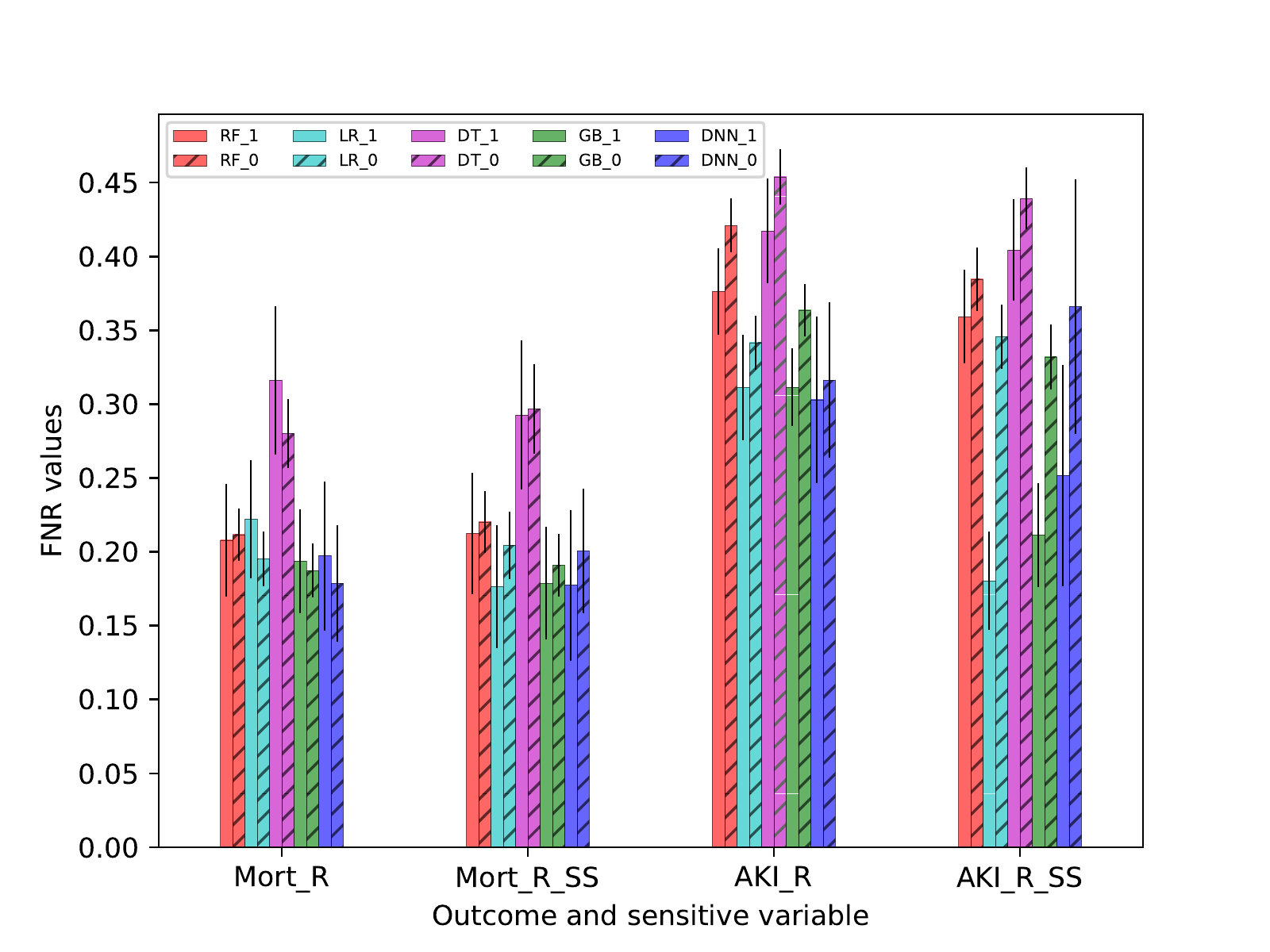}
    \end{subfigure}
    \vskip\baselineskip
    \begin{subfigure}[b]{0.8\textwidth}  
        \centering 
        \includegraphics[width=0.85\textwidth]{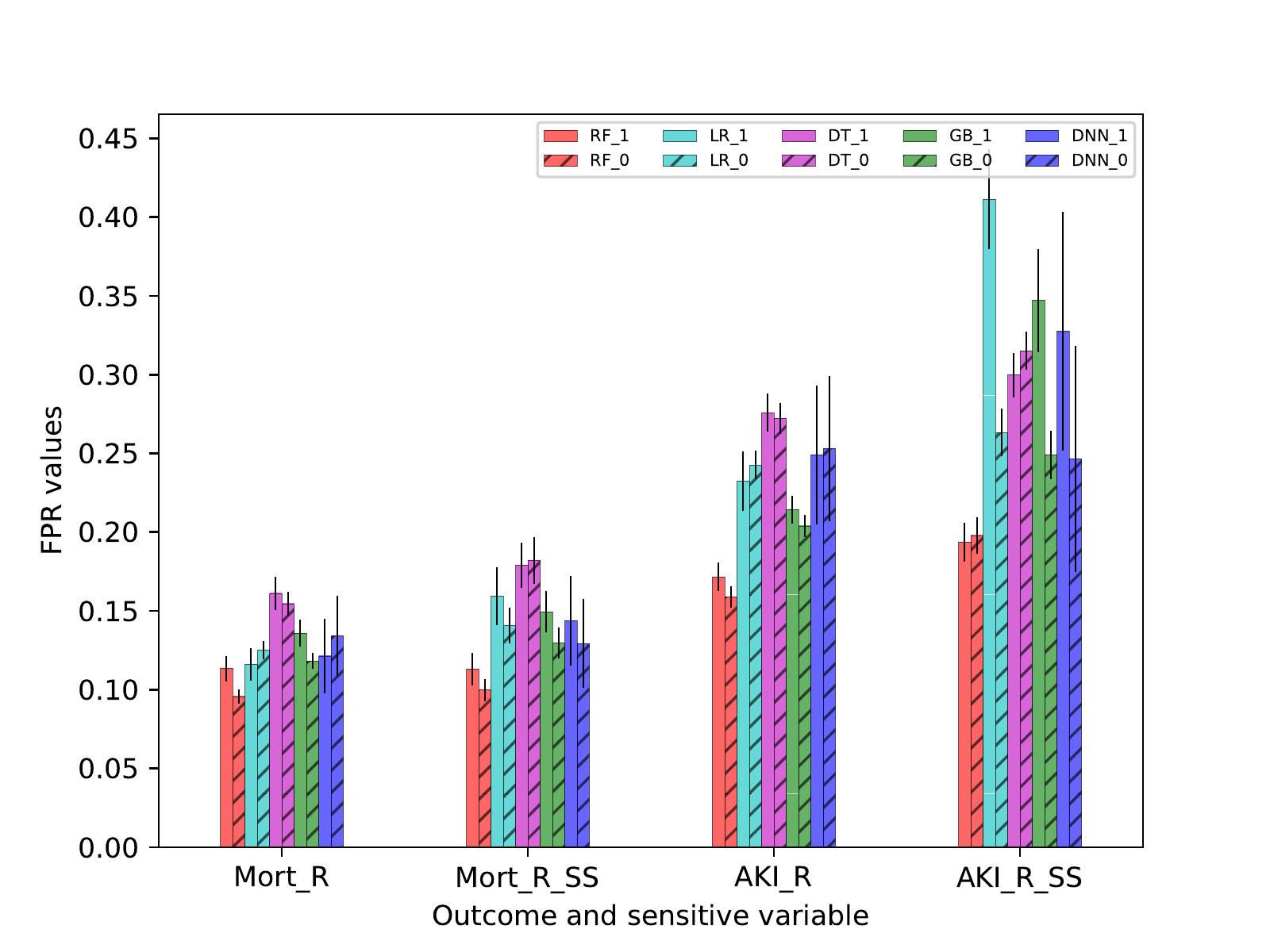}
    \end{subfigure}
        \caption{Figure comparing the performance measure $\in$ \{FNR, FPR\} values across subgroups of sensitive characteristic race (R) when the models were trained on down-sampled majority subgroup data for outcomes 30 day mortality (Mort) and AKI. Full and SS denotes training on full and down-sampled data.}
    \label{figapp: subs_models_diff_perf_FNR_FPR}
\end{figure*}

\begin{figure*}[h!]
    \centering
        \begin{subfigure}[b]{0.8\textwidth}
        \centering
        \includegraphics[width=1\textwidth]{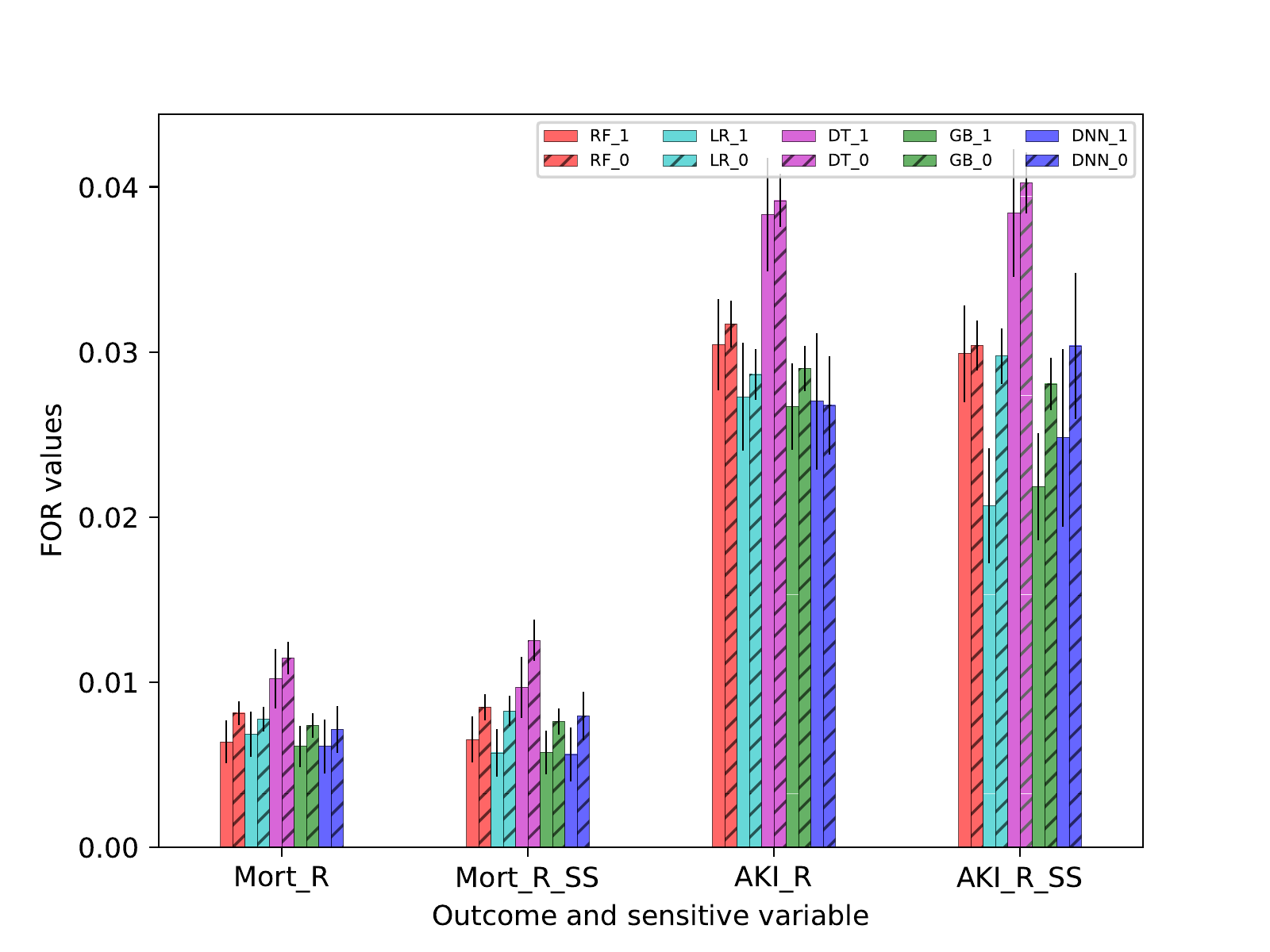}
    \end{subfigure}
    \vskip\baselineskip
    \begin{subfigure}[b]{0.8\textwidth}   
        \centering 
        \includegraphics[width=1\textwidth]{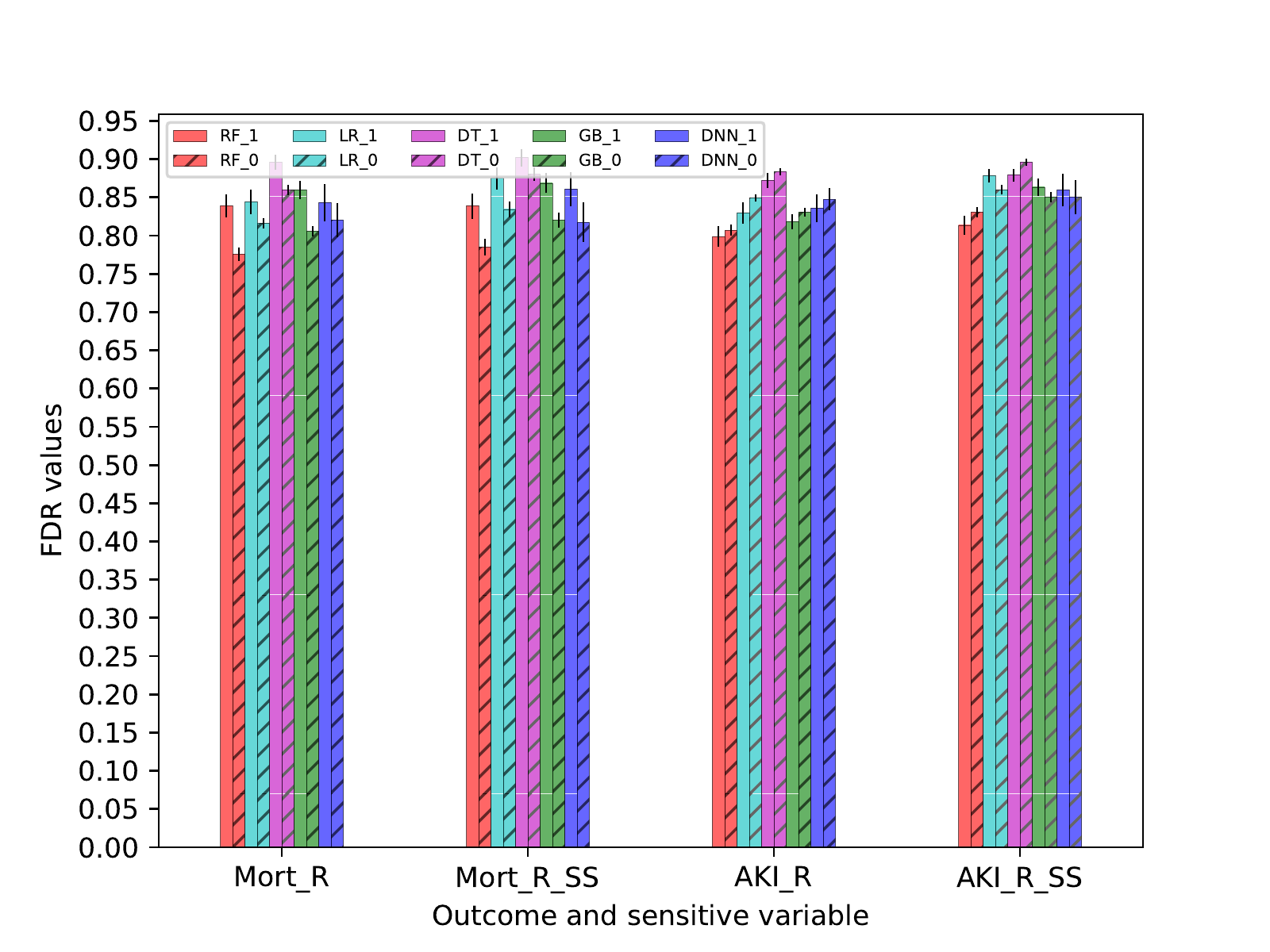}
    \end{subfigure}
    \caption{Figure comparing the performance measure $\in$ \{FOR, FDR\} values across subgroups of sensitive characteristic race (R) when the models were trained on down-sampled majority subgroup data for outcomes 30 day mortality (Mort) and AKI. Full and SS denotes training on full and down-sampled data.}
    \label{figapp: subs_models_diff_perf_FOR_FDR}
\end{figure*}

In Figure \ref{fig: subs_models_diff_perf} we show the effect of downsampling by race.
Because sex is nearly balanced, we do not show downsampling results.
To improve the readability of the figures, we show results only for two examplars: RF (because it was the best overall performing classifier, is highly flexible, and is commonly used) and LR (because it is the simplest model that is commonly used in practice, strongly averages effects of covariables across levels of Z, and for $n \ll m$ is relatively resistant to over-fitting ).
We observed that GB stumps and DNN (shown in SM Figures \ref{figapp: subs_models_diff_perf_FNR_FPR} and \ref{figapp: subs_models_diff_perf_FOR_FDR}) had similar results to RF, with GB showing somewhat exaggerated differences and DNN somewhat smaller.
We found that DT was overall a high-variance classifier; this probably relates to the complexity limitations related to our hyperparameter selections.

Interestingly, balancing the group sizes did not alleviate differences, and in several cases made them worse.
One might expect the downsampling to selectively reduce performance in the more flexible classifiers prone to overfitting, but logistic regression has the highest variability in performance.

\begin{figure*}[h!]
    \centering
    \begin{subfigure}[b]{0.475\textwidth}
        \centering
        \includegraphics[width=1\textwidth]{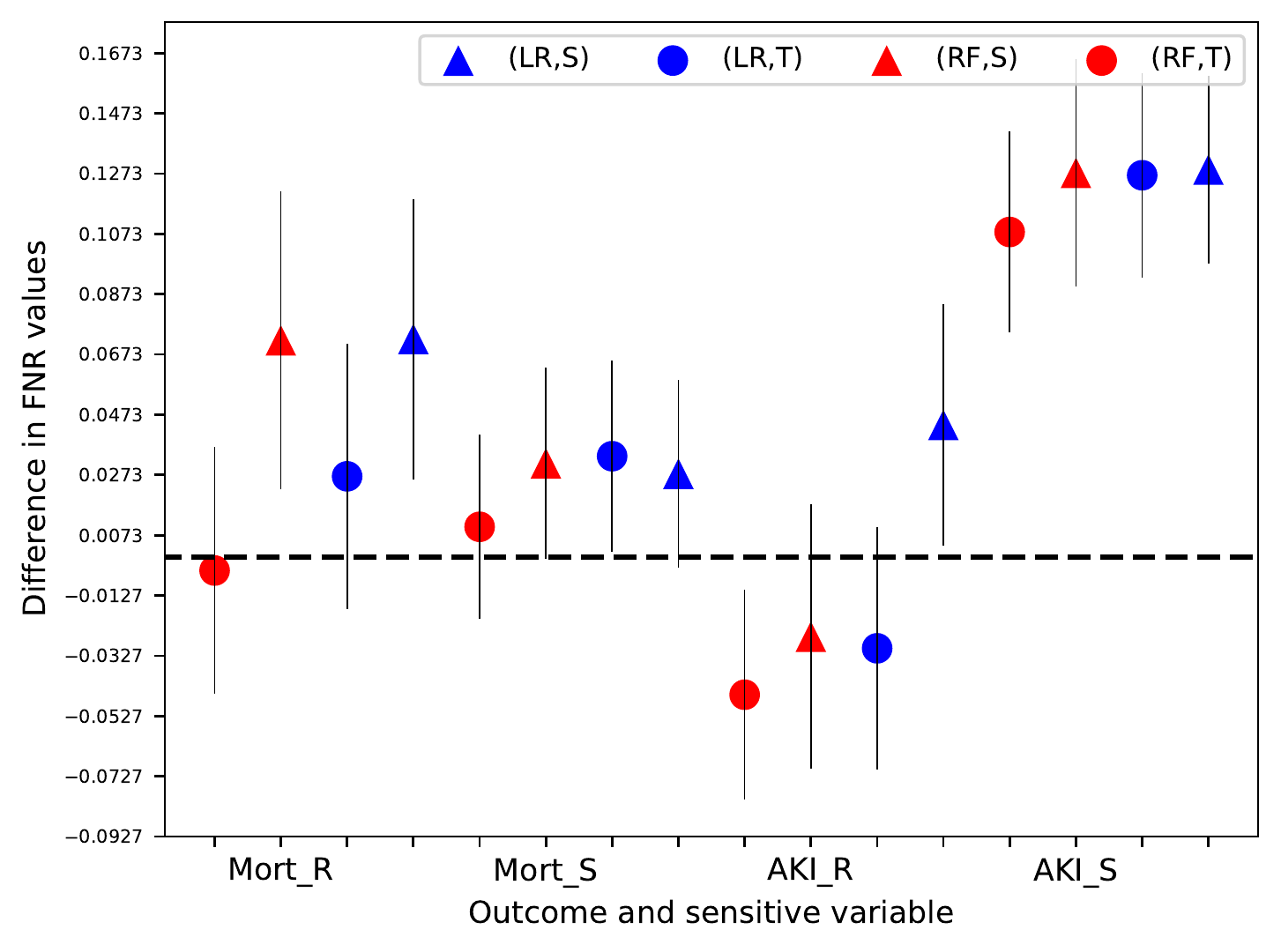}
    \end{subfigure}
    \hfill
    \begin{subfigure}[b]{0.475\textwidth}  
        \centering 
        \includegraphics[width=1\textwidth]{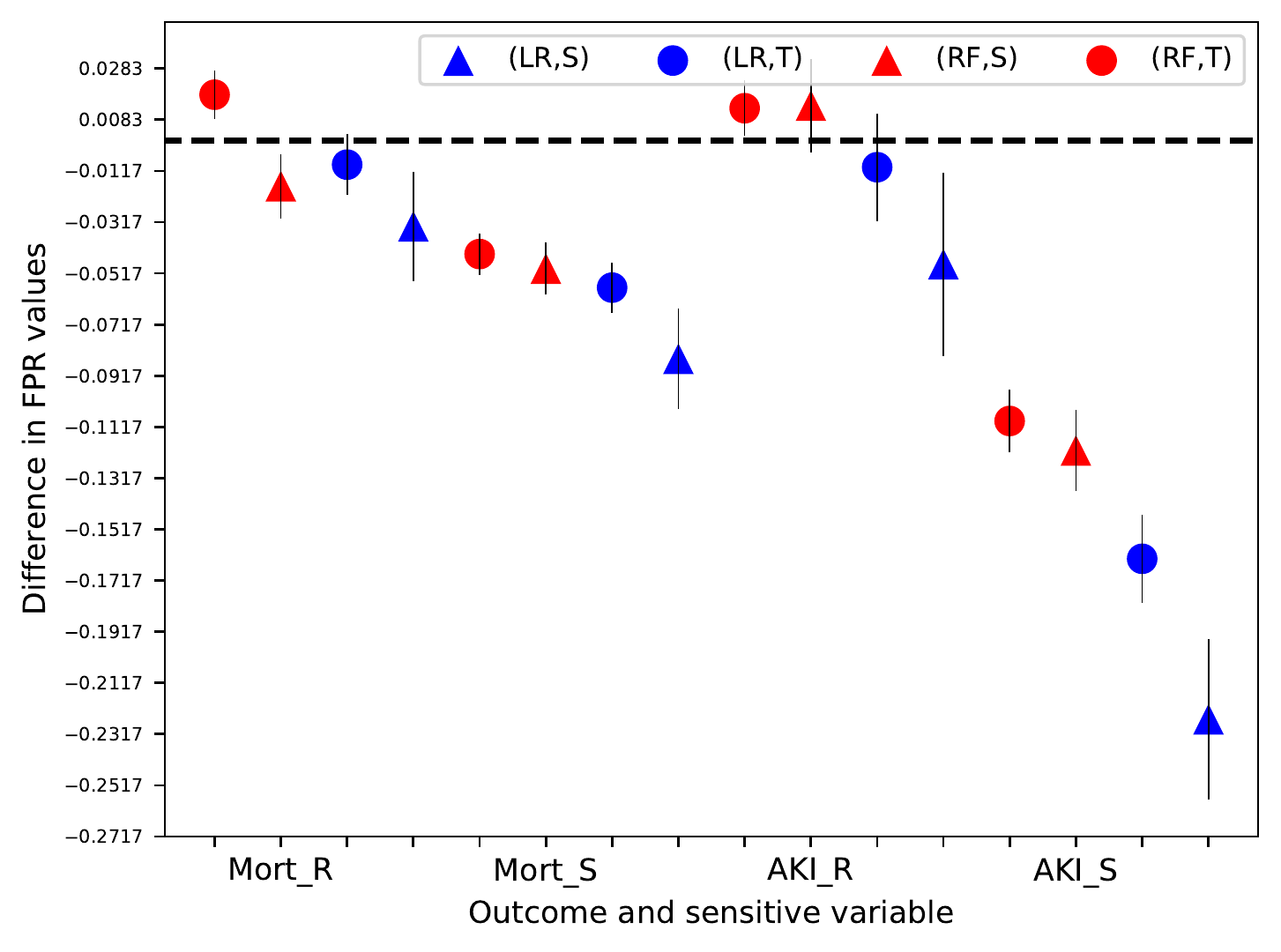}
    \end{subfigure}
    \caption{Each panel in the figure depicts the difference in performance measure $\in$ {FNR, FPR} for the classifiers $\in$ {RF, LR} marked by color across sensitive characteristic Race (R) or Sex (S) subgroups for the outcomes 30 day mortality (Mort) and AKI. Markers $\Delta$ and $\circ$ denote whether the classifiers were trained separately (S) on subgroups of a particular value of sensitive characteristic or both subgroup data was combined together (T) for training. Actual values of FNR, FPR along with FOR and FDR for all classifiers can be found in SM Figure \ref{figapp: per_on_sep_groups_FPR_FNR} and \ref{figapp: per_on_sep_groups_FOR_FDR}.}
    \label{fig: per_on_sep_groups}
\end{figure*}

Next, we tested the performance of combined learner on the subgroups from the test datasets and compared it to the performance of the individual learners for the respective subgroups. 
The intention behind this experiment is to see if differential performance results from applying patterns across the subgroup; if so the disparity will improve if the learner is trained only on the subgroup dataset. 
We present the results of this experiment in Figure \ref{fig: per_on_sep_groups}. 
Disparities are somewhat improved by training separate classifiers.
There is no consistent difference between the model with interactions (RF) and the model without (LR).

\subsection{Group Differences and Performance} \label{subsec: direct_disc}
Next, we examine the role of correlates of the sensitive feature in explaining performance differences.
Similar to the case of bias, we examine if the difference in performance is removed by scrambling or omitting the sensitive feature (which would suggest that it relates to the value of the sensitive feature tagging unobserved characteristics) and by evaluating in a propensity-matched set (which would suggest it is the result of differences in observed characteristics). 

\begin{figure*}[h!]
    \centering
    \begin{subfigure}[b]{0.475\textwidth}
        \centering
        \includegraphics[width=1\textwidth]{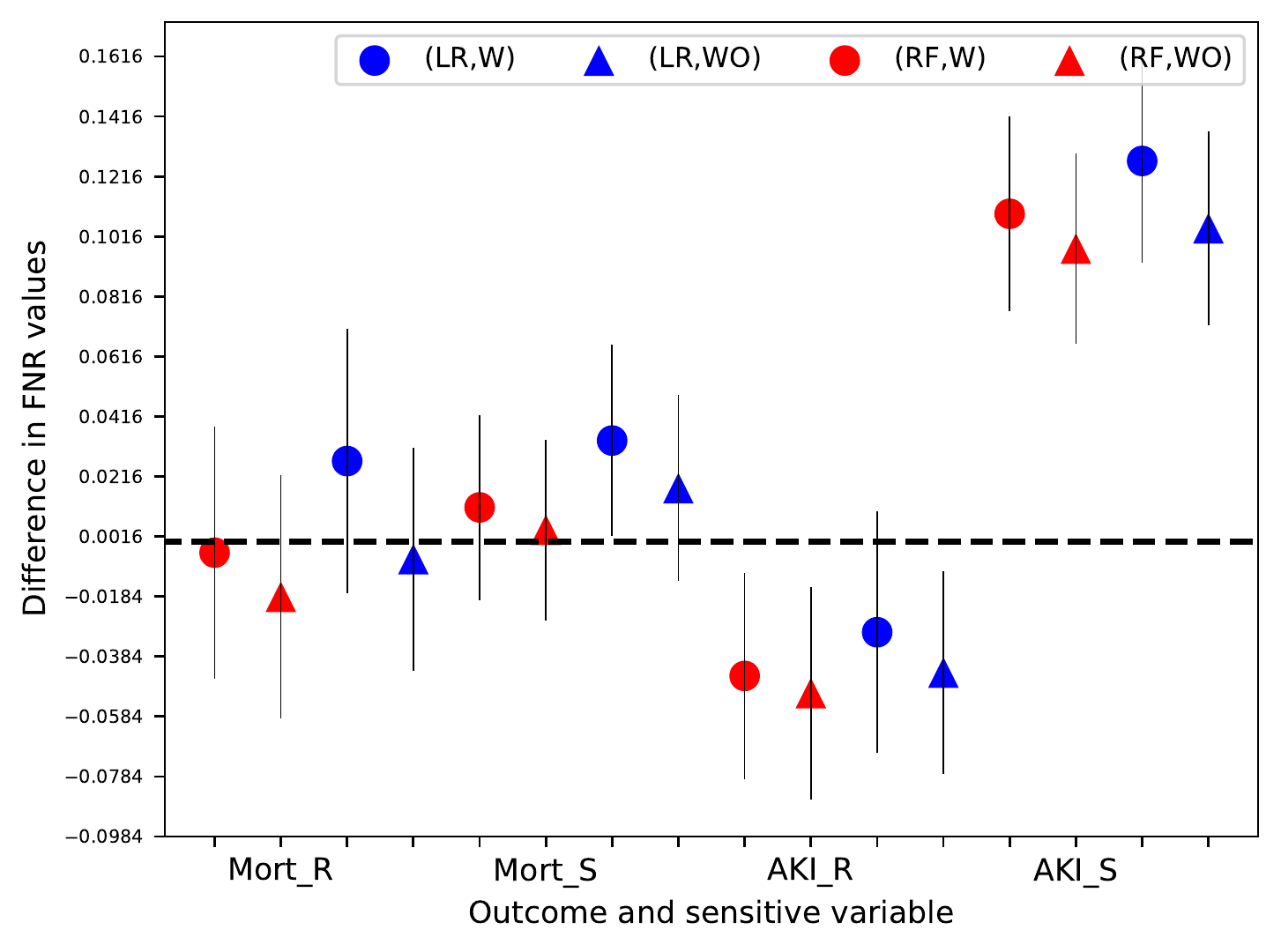}
    \end{subfigure}
    \hfill
    \begin{subfigure}[b]{0.475\textwidth}  
        \centering 
        \includegraphics[width=1\textwidth]{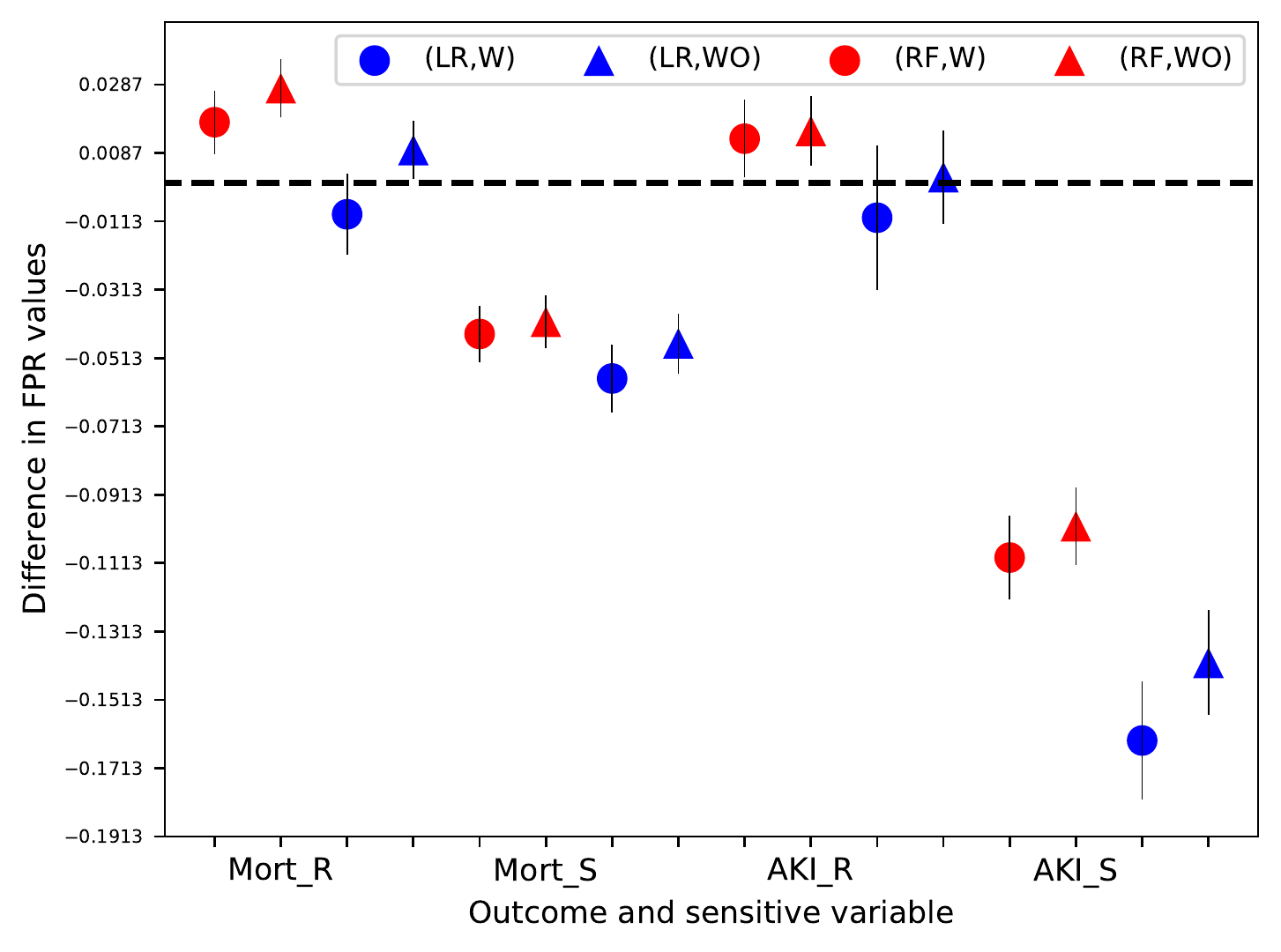}
    \end{subfigure}
    \caption{Comparison of differences in performance measures $\in$\{FNR, FPR\} for the cases when the sensitive variable was used (W) or not (WO) while training across 
    Race (R) and Sex (S) for the outcomes 30 day mortality (Mort) and AKI. Similar pattern was observed for the performance measures FDR and FOR. As can be seen, FNR differences decreased but the FPR differences increased even though the magnitude is very small. Overall, removing the sensitive characteristic doesn't decrease the bias. Actual values of FNR, FPR along with FOR and FDR for all classifiers can be found in SM Figure \ref{figapp: basic_WO_models_diff_perf_FPR_FNR} and \ref{figapp: basic_WO_models_diff_perf_FOR_FDR}.}
    \label{fig: basic_WO_models_diff_perf}
\end{figure*}

\begin{figure*}[h!]
    \centering
    \begin{subfigure}[b]{0.475\textwidth}
        \centering
        \includegraphics[width=1\textwidth]{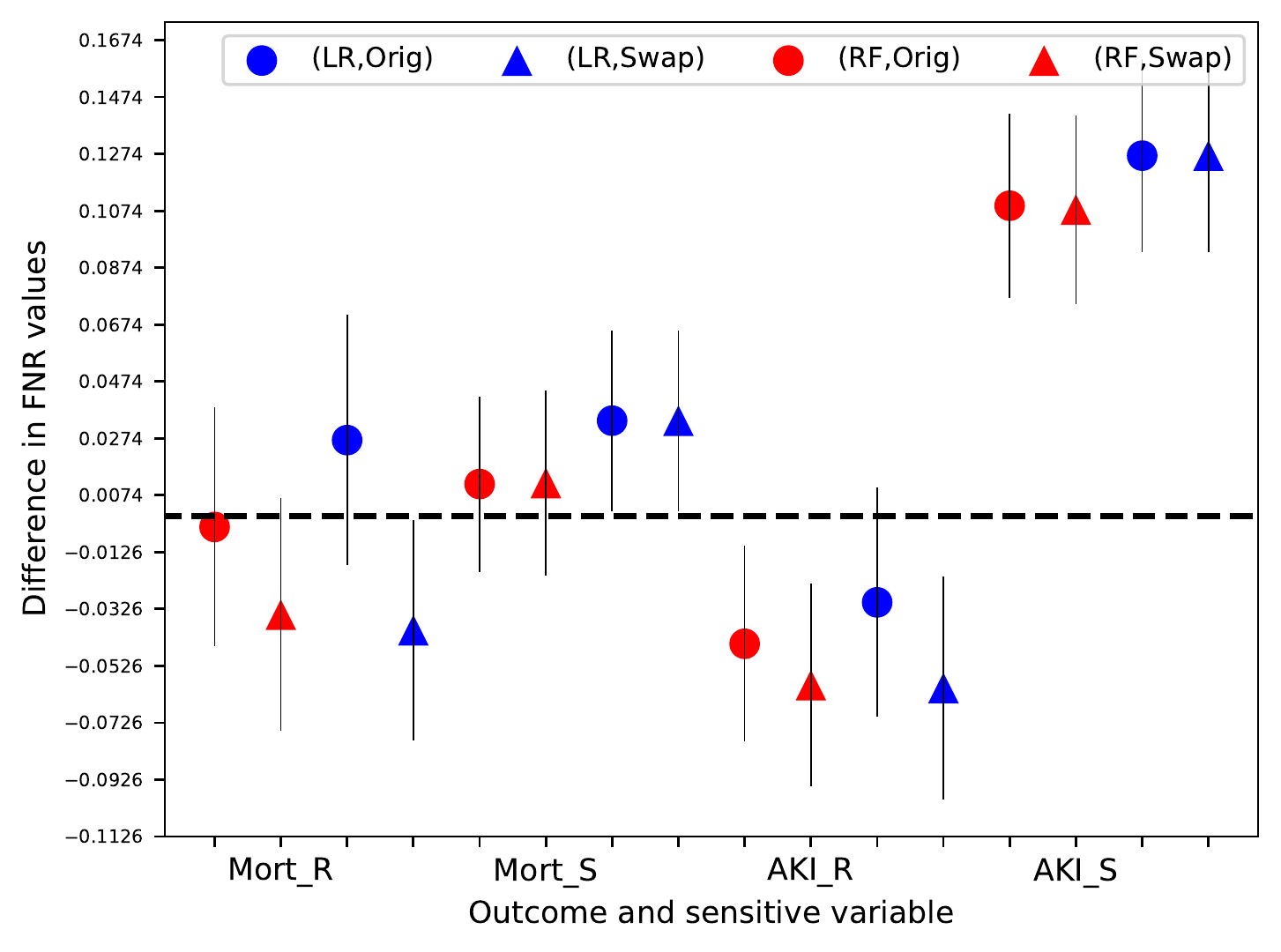}
    \end{subfigure}
    \hfill
    \begin{subfigure}[b]{0.475\textwidth}  
        \centering 
        \includegraphics[width=1\textwidth]{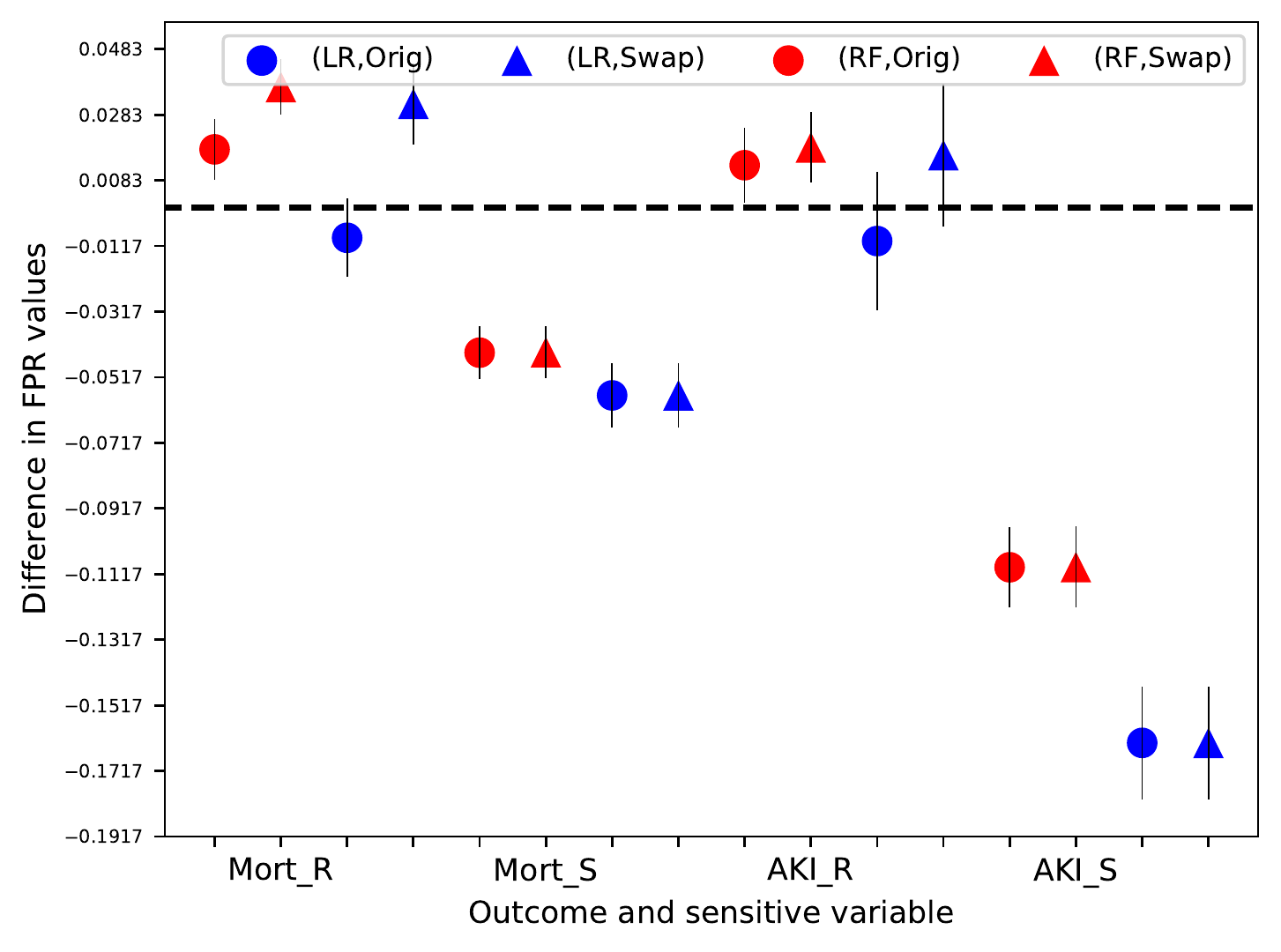}
    \end{subfigure}
    \caption{Comparison of difference in performance measures $\in$ \{FNR, FPR\} of classifiers with true value of sensitive attribute (Orig) and when the value of sensitive attribute is inverted/swapped (Swap) for Race (R) and Sex (S) for the outcomes 30 day mortality (Mort) and AKI.  }
    \label{fig: per_after_swapping}
\end{figure*}

To further examine for potential differences due to features correlated with the sensitive characteristic, we stratify on several major features.
We also address a commonly made argument for explaining the differences in performance across subgroups: "the differential performance is due to the inclusion of the sensitive features while training". 
Figure \ref{fig: basic_WO_models_diff_perf} displays the results with and without the sensitive feature included in training.
The differential performance is not affected. 
 Figure \ref{fig: per_after_swapping} displays when the value of sensitive characteristic was inverted. In particular, for both the outcomes and sex as the sensitive characteristic there is no change in differential performance. However, the differences exist both for FNR and FPR. This suggests that there is more to the differences than ``direct discrimination''.

\begin{figure*}[h!]
    \centering
    \begin{subfigure}[b]{0.475\textwidth}
        \centering
        \includegraphics[width=1\textwidth]{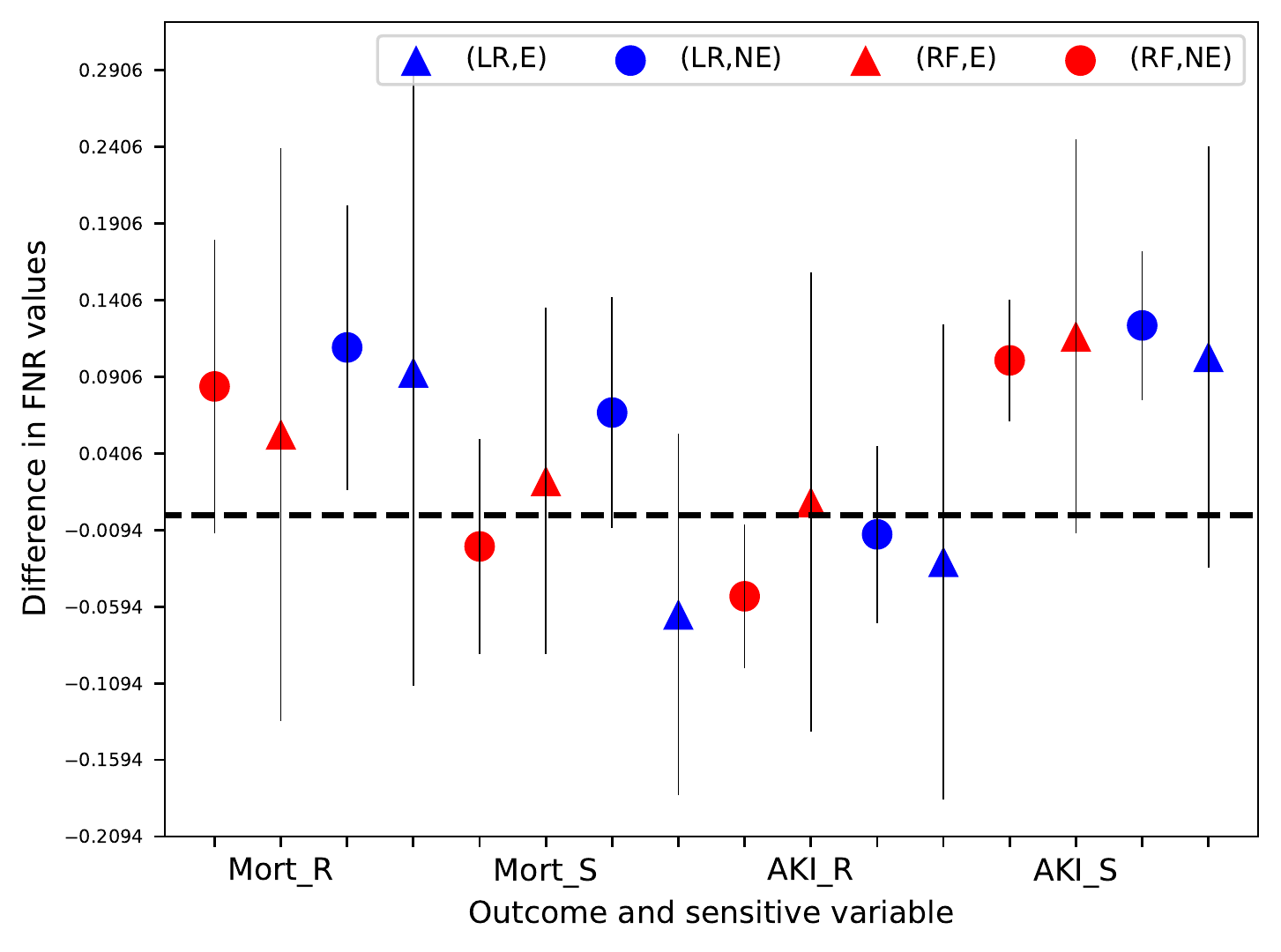}
    \end{subfigure}
    \hfill
    \begin{subfigure}[b]{0.475\textwidth}  
        \centering 
        \includegraphics[width=1\textwidth]{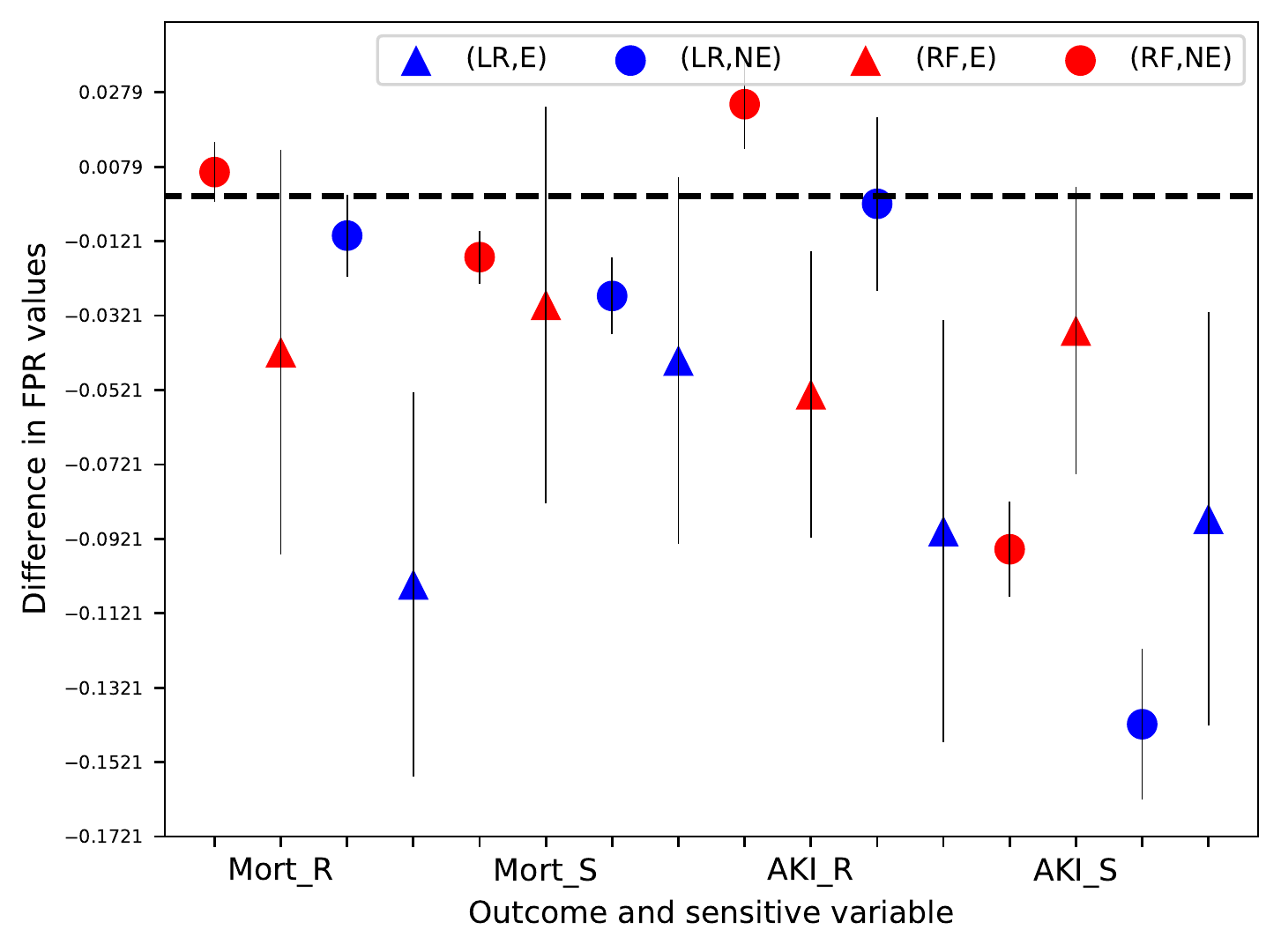}
    \end{subfigure}
    \caption{Comparison of differences in performance measures $\in$ \{FNR, FPR\} on the group ASA2\_NE and ASA2\_E (patients having systematic disease) across the values taken by sensitive characteristics Race (R) and Sex (S) for the outcomes 30 day mortality (Mort) and AKI. E and NE denotes group categorization in ASA emergency case or not. Actual FNR, FPR values for all classifiers can be found in SM Figure \ref{figapp: strat_ASA2_FNFR_FPR} and \ref{figapp: strat_ASA2_FOR_FDR}.}
    \label{fig: strat_ASA2}
\end{figure*}

\begin{figure*}[h!]
    \centering
    \begin{subfigure}[b]{0.475\textwidth}
        \centering
        \includegraphics[width=1\textwidth]{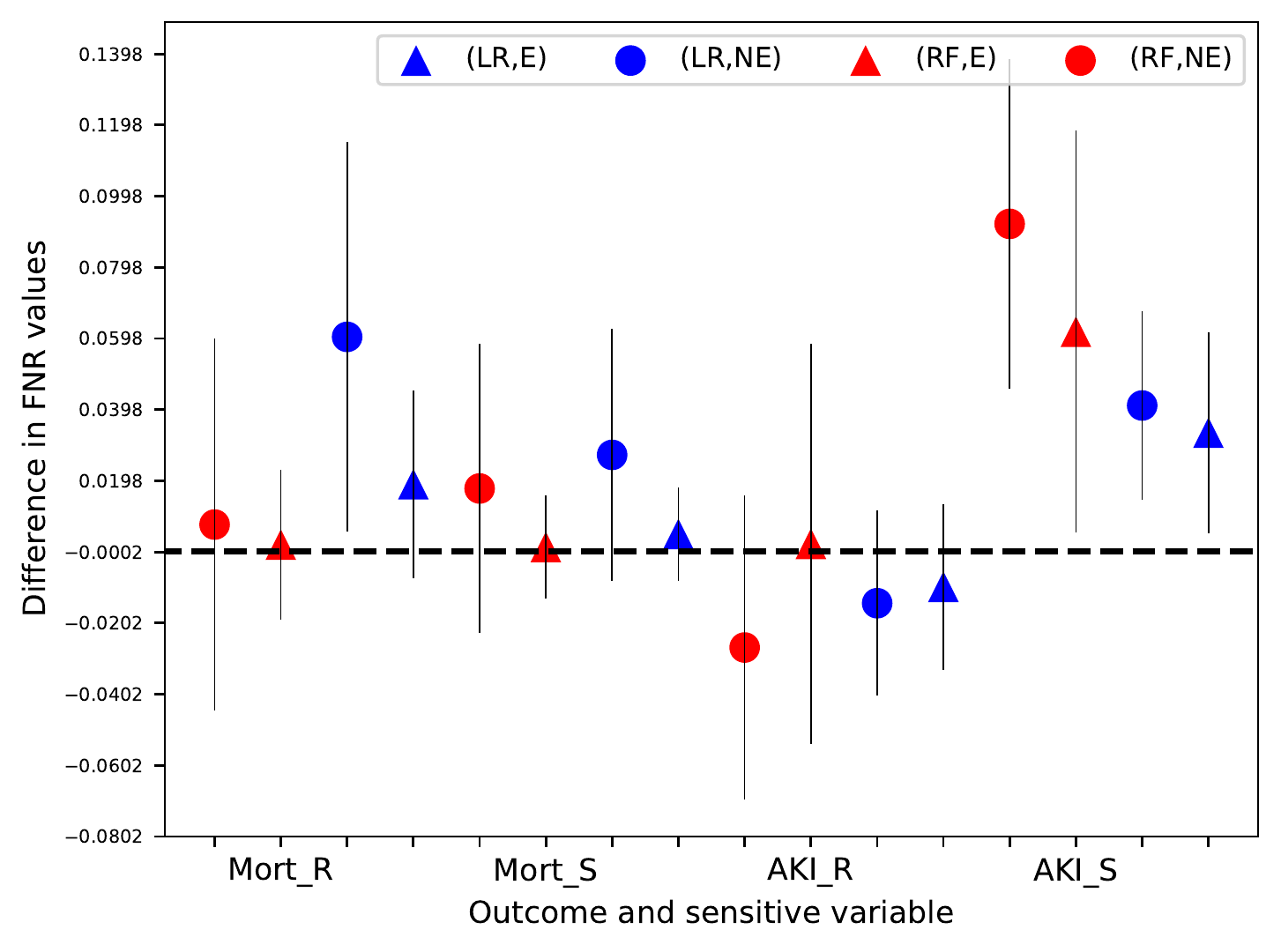}
    \end{subfigure}
    \hfill
    \begin{subfigure}[b]{0.475\textwidth}  
        \centering 
        \includegraphics[width=1\textwidth]{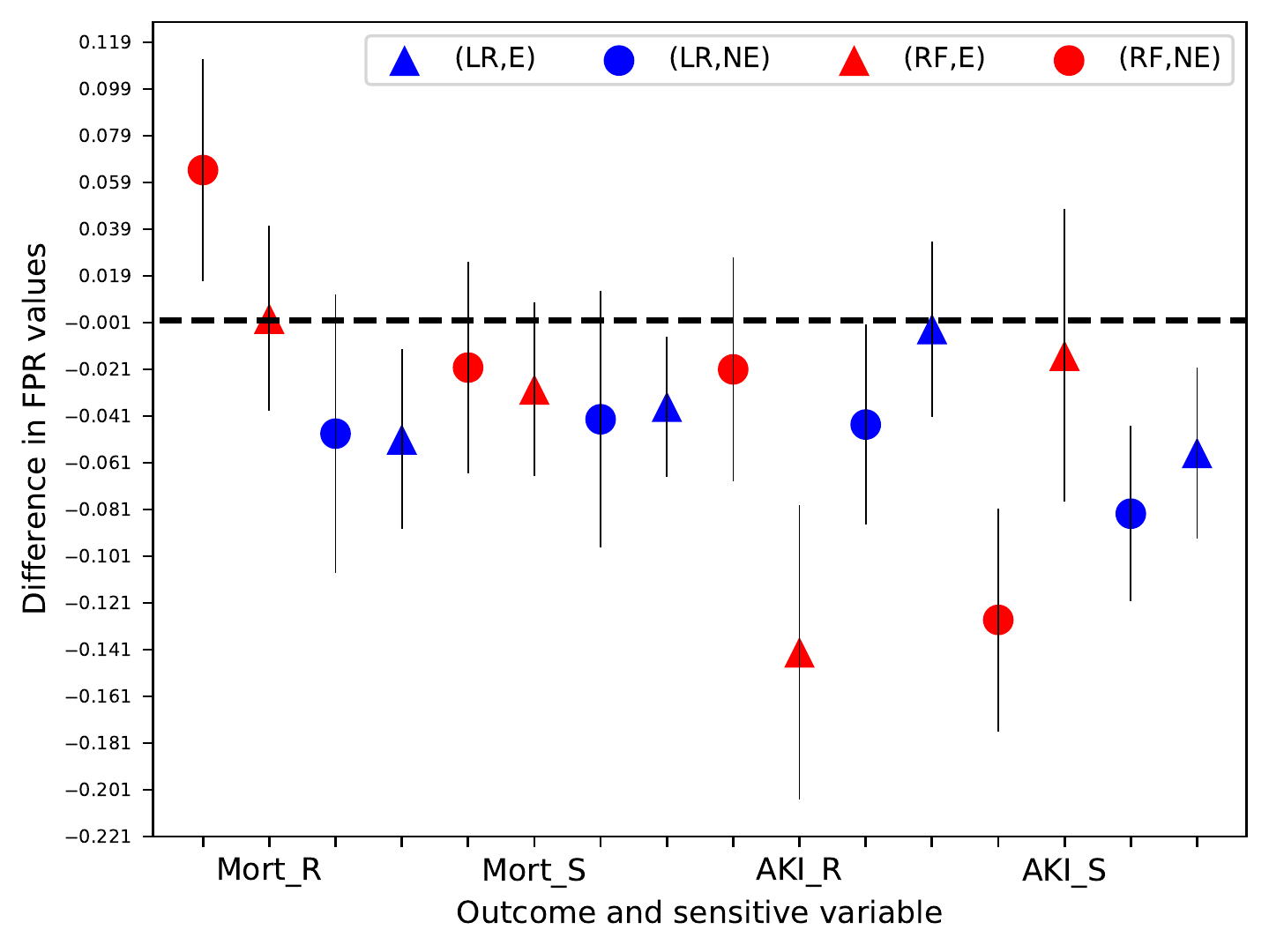}
    \end{subfigure}
    \caption{Comparison of differences in performance measures $\in$ \{FNR, FPR\} on the group ASA3\_NE and ASA3\_E (patients having constant threat to life) across the values taken by sensitive characteristics Race (R) and Sex (S) for the outcomes 30 day mortality (Mort) and AKI. E and NE denotes group categorization in ASA emergency case or not. Actual FNR, FPR values for all classifiers can be found in SM Figure \ref{figapp: strat_ASA3_FNR_FPR} and \ref{figapp: strat_ASA3_FOR_FDR}.}
    \label{fig: strat_ASA3}
\end{figure*}

To detect differences in performance related to covariables, Figure \ref{fig: strat_ASA2} and \ref{fig: strat_ASA3} illustrates the results stratifying the accuracy measures on ASA-PS, a global subjective assessment of health performed by the anesthesiologist. \cite{aso2014asa} 
 Our initial experiments suggested that ASA physical status is one of the most important features for predicting the outcome 30 day mortality and AKI. 
Domain experts also agreed that based on the existing literature this feature was likely to be very influential and absorb much of the information of specific comorbidities.
A patient can be assigned to any of the six categories: ASA I (healthy), ASA II (mild systematic disease), ASA III (severe systematic disease), ASA IV (severe systematic disease that has constant threat to life), ASA V (surgery is necessary for survival), ASA VI (brain dead organ donor) \cite{aso2014asa}. 
The "E" modifier refers to the situation where the delay in surgery would increase the threat to patient's life.
To simplify presentation, we merged several categories using input from clinical domain experts.
ASA1\_NE (ASA1\_E) has patients that were classified as healthy and non-emergency (emergency). 
ASA2\_NE (ASA2\_E) combines the patients that were classified as ASA II or ASA III, i.e., have systematic disease and non-emergency (emergency). 
ASA3\_NE (ASA3\_E) combines the patients that were classified as ASA IV or ASA V, i.e., have a constant threat to life and non-emergency (emergency).


As the true outcome rate for ASA category 1 (i.e., healthy patients) is 0 and 0.001 for 30 day mortality and AKI respectively, the comparison of FNR and FPR across two groups of sensitive characteristics is not conclusive. For the other two cases, the results are presented in Figure \ref{fig: strat_ASA2} and \ref{fig: strat_ASA3}. For patients belonging to ASA2\_NE and ASA2\_E category and when the outcome to be predicted is 30 day Mortality, the FNR value is consistently higher for black race people across all algorithms. FPR plot further supports this where the FPR value is higher for white race people when the patients were operated as an emergency case. For patients from ASA3\_NE and ASA3\_E category, there were enough positive outcome examples for the classifier to learn and hence, the FNR value is generally low. However, one can still observe some differences between the FNR value for two subgroups of sensitive characteristic race. 

\begin{figure*}[h!]
    \centering
    \begin{subfigure}[b]{0.475\textwidth}
        \centering
        \includegraphics[width=1\textwidth]{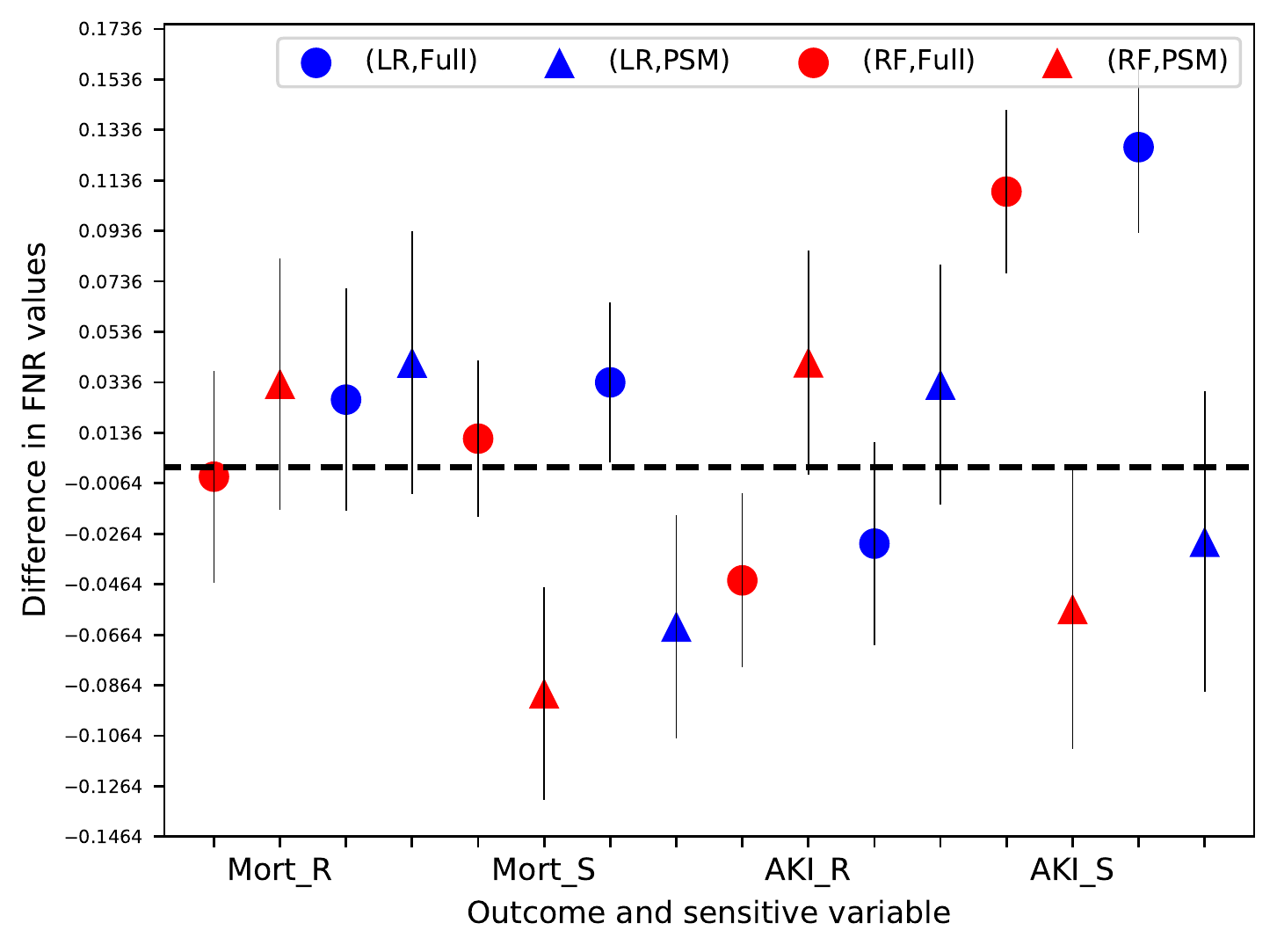}
    \end{subfigure}
    \hfill
    \begin{subfigure}[b]{0.475\textwidth}  
        \centering 
        \includegraphics[width=1\textwidth]{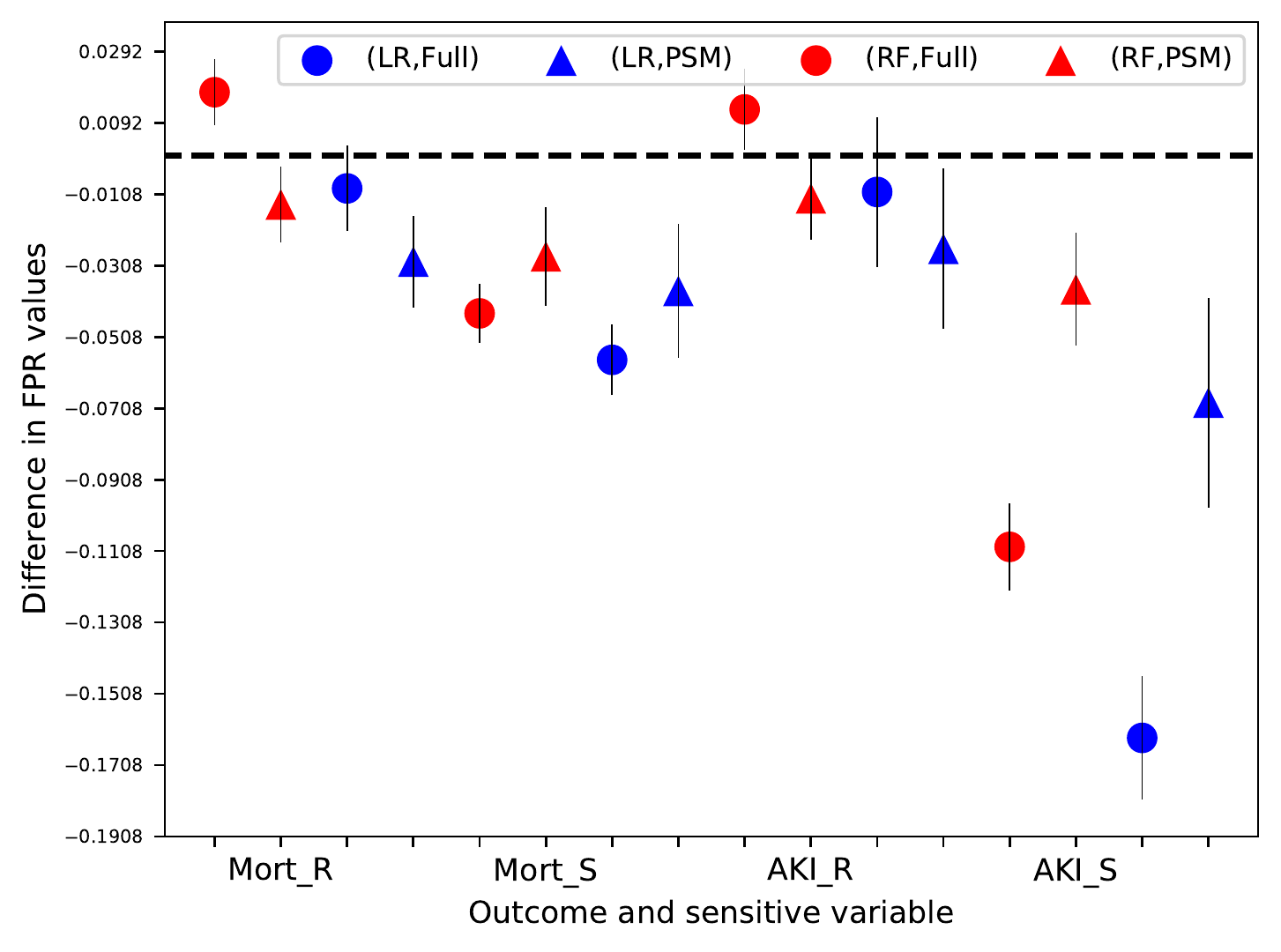}
    \end{subfigure}
    \caption{Comparison of differences in performance measures $\in$ \{FNR, FPR\} on complete test set (Full) and  propensity matched test set (PSM) across the subgroups of sensitive characteristics Race (R) and Sex (S) for the outcomes 30 day mortality (Mort) and AKI. Actual FNR, FPR along with FOR and FDR values for all classifiers can be found in SM Figure \ref{figapp: per_on_PSM_set_FNR_FPR} and \ref{figapp: per_on_PSM_set_FOR_FDR}.}
    \label{fig: per_on_PSM_set}
\end{figure*}

The results of various classifiers on the Propensity matched samples across the subgroups of sensitive attributes are presented in Figure \ref{fig: per_on_PSM_set}. 
Differential performance was affected a great deal (especially across sex), but not in a consistent way; occasionally the sign is even reversed.
Collectively, these suggest that differences in covariables explain only part of the disparity in predictive accuracy.

\section{Passive approaches for bias reduction} \label{sec: passive_app}
So far we have stratified model performance on pre-specified sensitive characteristics.
However, this may obscure other features which ``stand in'' for a sensitive characteristic in an unexpected way.
Additionally, interactions with other features may reflect important deficiencies in the model.
Ideally, an adaptive approach would answer the question ``which patients benefit from the use of the model?'' in a manner transparent to the end user, similar to the ``model card'' approach. \cite{mitchell2019modelcards}
This would allow the decision maker to decide whether to trust the model (and intervene accordingly) or to take potential model deficiencies into account in the case under scrutiny.
In this section, we propose an approach that helps us understand how useful the full model prediction have been in comparison to a very basic model. 
Here, the ``full model'' is the one that uses all the pre-operative data and the ``basic model'' is one that uses only type of surgery, age, race, and sex. 
The basic model (rather than a null one) is employed to acknowledge that clinicians often have a learned heuristic about the importance of these common, easily observed factors.
The utility for every patient is defined in a manner analogous to the ``equalized odds'' criterion \cite{rajkomar_ensuring_2018}.
Unlike the binary cross entropy loss, which is defined as $-(y\log(\hat(p))+(1-y)\log(1-\hat{p}))$ where $y\in \{0,1\}$ is true label and $\hat{p}\in (0,1)$ is the predicted in-class probability, we focus on the improvement of the absolute probability of a correct prediction. 
We define the weighted utility to be
\begin{equation} \label{eq: Indi_utility}
    IU = w_1 y \hat{p}+w_2 (1-y) (1-\hat{p}).
\end{equation}
The weights $(w_1,w_2)$ correspond to the relative importance of correct positive class and  correct negative class predictions, respectively.
This is a re-weighted version of the mean absolute error, which is an easily interpreted quantity by users, in contrast to cross-entropy.
Let $IU_{Mfull}$ and $IU_{Mbasic}$ denote the performance of full model and basic model on a particular patient. 
As we are interested in the improvement in utility obtained by the use of full model, we define $IU_{diff} = IU_{Mfull} - IU_{Mbasic}$. 
To identify which patients experience increased or decreased utility, we consider a model that regresses $IU_{diff}$ on available features. 
Because it has a high level of transparency and forms natural groups, we train a decision tree on $IU_{diff}$ in the test dataset. 

\begin{figure*}[!ht]
    \centering
    \begin{subfigure}[b]{0.475\textwidth}
        \centering
        \includegraphics[width=1.1\textwidth]{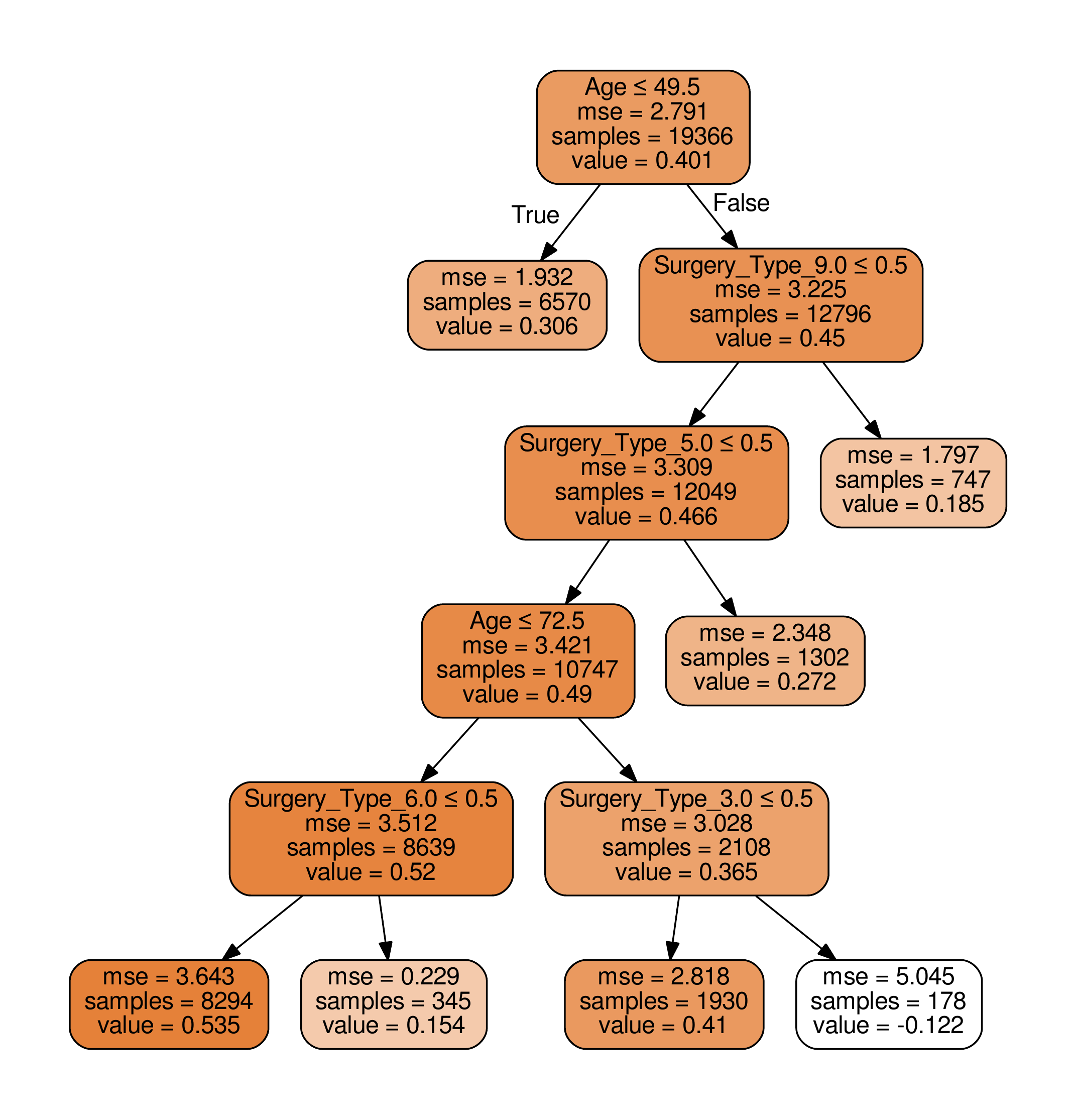}
    \end{subfigure}
    \hfill
    \begin{subfigure}[b]{0.475\textwidth}  
        \centering 
        \includegraphics[width=1.1\textwidth]{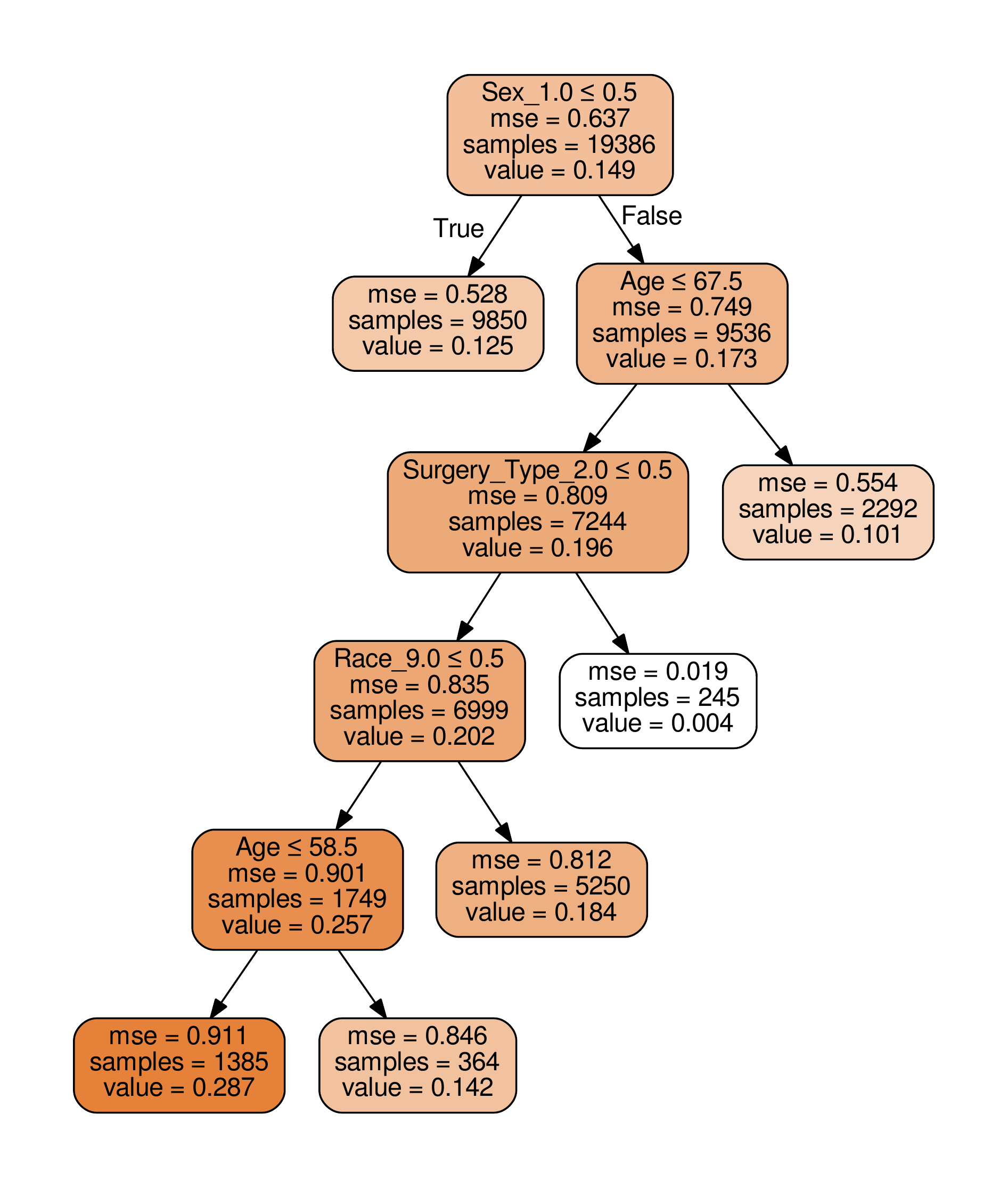}
    \end{subfigure}
    \caption{Left (Right) panel has 30 day Mortality (AKI)  outcome with utility computed for an LR classifier. 'value' at each node denotes the improvement in utility obtained by using a classifier trained on all features in comparison to a classifier trained on only surgery type, age, sex and race. Minimal cost complexity pruning with the complexity parameter $\alpha$ taking the $5th$ highest value from the $\alpha$ path. Minimum number of sample to split and minimum number of samples for a leaf node were set to be 50. The mapping between surgery type number, sex and race is provided in Table \ref{tab: dict_surgerytype}.}
    \label{fig: decision_tree_utility}
\end{figure*}

To be consistent with our initial cost value for negative and positive class mis-classification, we have taken $(w_1,w_2)$ to be $(25,1)$ and $(14,1)$ for 30 day mortality and AKI respectively. 
An example of such a decision tree is provided in Figure \ref{fig: decision_tree_utility}, which uses LR as the ``full model'' and ``base model'' classifier. 
Because LR tends to produce nearly calibrated outputs, this avoids the need to include a calibration step mapping outputs to probabilities.
These trees demonstrate the transparency and interpetability of the scheme for the users as they can easily read this for a patient and decide how much to trust the risk predictions. 
For example, patients between 50 and 72 benefit more than other age groups in mortality prediction, with several surgical specialties benefiting less than others.
In particular, a subset of thoracic surgery patients experience \textit{decreased} utility compared to the minimal model.
Interesting, although Figure \ref{fig: basic_models_diff_perf } would suggest that sex should be an important variable, when comparing to a base model including sex, it is not selected as a node.
In contrast, the AKI utility decision tree includes race as a (large) predictor of utility and sex as a relatively smaller one; Figure \ref{fig: basic_models_diff_perf } would have suggested the opposite.

\begin{table*}[!htbp]
\centering
\setlength{\tabcolsep}{1.6pt}
\renewcommand{\arraystretch}{0.9}
{ \small
\begin{tabular}{|c|c|c|c|c|c|c|c|}
\hline
\textbf{ST\_0.0} & Gynecology & \textbf{ST\_5.0} & \begin{tabular}[c]{@{}c@{}}Orthopedic/ \\ Orthopedic spine\end{tabular} & \textbf{ST\_10.0} & Cardiovascular & \textbf{Sex\_1.0} & Male \\ \hline
\textbf{ST\_1.0} & \begin{tabular}[c]{@{}c@{}}Acute Critical care surgery, colorectal, \\ hepatobiliary, minimally invasive surgery,\\ pediatric, trauma, general surgery\end{tabular} & \textbf{ST\_6.0} & \begin{tabular}[c]{@{}c@{}}Neurosurgery/ \\ Neurosurgery spine\end{tabular} & \textbf{ST\_11.0} & \begin{tabular}[c]{@{}c@{}}Anesthesiology, minor procedures, gastroenterology,\\ oral/maxillofacial, pain management, pulmonary, \\ dental, oncology, transplant, hepatology\end{tabular} & \textbf{Sex\_2.0} & Female \\ \hline
\textbf{ST\_2.0} & Otolaryngology & \textbf{ST\_7.0} & Plastic & \textbf{ST\_12.0} & Opthalomology & \textbf{Race\_7.0} & Black \\ \hline
\textbf{\begin{tabular}[c]{@{}c@{}}ST\_3.0\\ ST\_14.0\end{tabular}} & Cardiothoracic & \textbf{ST\_8.0} & Transplant & \textbf{ST\_13.0} & Radiation oncology & \textbf{Race\_9.0} & White \\ \hline
\textbf{ST\_4.0} & Vascular & \textbf{ST\_9.0} & Urology & \textbf{} &  & \textbf{} &  \\ \hline
\end{tabular}}
\caption{Data dictionary (for sex, race and surgery type) to be used while reading the decision trees in Figure \ref{fig: decision_tree_utility}. `Surgery\_Type\_*' has been shortened to `ST\_*' in the table where $*$ denotes the number.}
\label{tab: dict_surgerytype}
\end{table*}

\section{Discussion}

Detection of algorithmic bias has become a major concern in the application of predictive analytics to clinical data.
Importantly, we focus on the direct results of the algorithm and not biased interpretation or selective use of the outputs resulting from implicit bias within the user \cite{van_ryn_impact_2011}.
This concern is especially worrisome in the case of relatively opaque classifiers such as DNN and GBDT and when covariables may be surrogates for the sensitive characteristics.
This setting can be difficult to distinguish from the case that real differences in covariables explain important differences in predictions.

We step through a recommended procedure of comparing classifier output and performance across subgroups, testing for direct use of known sensitive characteristics,  pattern-transfer effects, and sample-size effects.
We also check for surrogates of the sensitive characteristic by use of propensity scores, and evaluate performance differences among patients with otherwise similar features.
We sanity-check these results by using domain experts to identify key covariables and stratify within those covariables.
Finally, we illustrate the use of individual level predicted utility as target for grouping algorithms to identify patients who gain much or little from the use of the algorithm.
Although some classifiers (such as LR and RF) have in-sample estimates of uncertainty, by using test sample predictions we can apply the method generically.

In our example application, we find large differences in performance in AKI prediction across sexes, and smaller differences across race.
None of our checks fully eliminate the sex differences, and our final utility-grouping approach identifies race and age as major drivers of algorithmic utility.
What residual sources of unfairness could explain these differences?
First, unmeasured variation in other variables is always a potential source of variance.
For example, there may be more variation in the surgeries performed on men than women above that is captured by the ``surgery type'' variable in the dataset.
Our dataset also contains very little social and economic information; these sources of differences could be more variable among men.
This is consistent with the shared pattern across mortality and AKI in the propensity matched set.
It is also well appreciated that historical epidemiology in the US struggled to adequately include participants other than white men, and so risk factors sought out in clinic may be those most relevant to this group. \cite{konkel_racial_2015,murthy_participation_2004}
We hope that by identifying the differential value of the available information, targeted improvement to data collection, algorithm development, and user education can be made to avoid worsening disparities with these new tools.


\bibliographystyle{ACM-Reference-Format}
\bibliography{sample-base}
\newpage
\onecolumn
\begin{center}
   {\LARGE \textbf{Supplementary material for} \\
    \textbf{(Un)fairness in Post-operative Complication Prediction Models}\\}
\end{center}


\begin{figure*}[h!]
    \centering
    \begin{subfigure}[b]{0.9\textwidth}
        \centering
        \includegraphics[width=20cm, height =11cm]{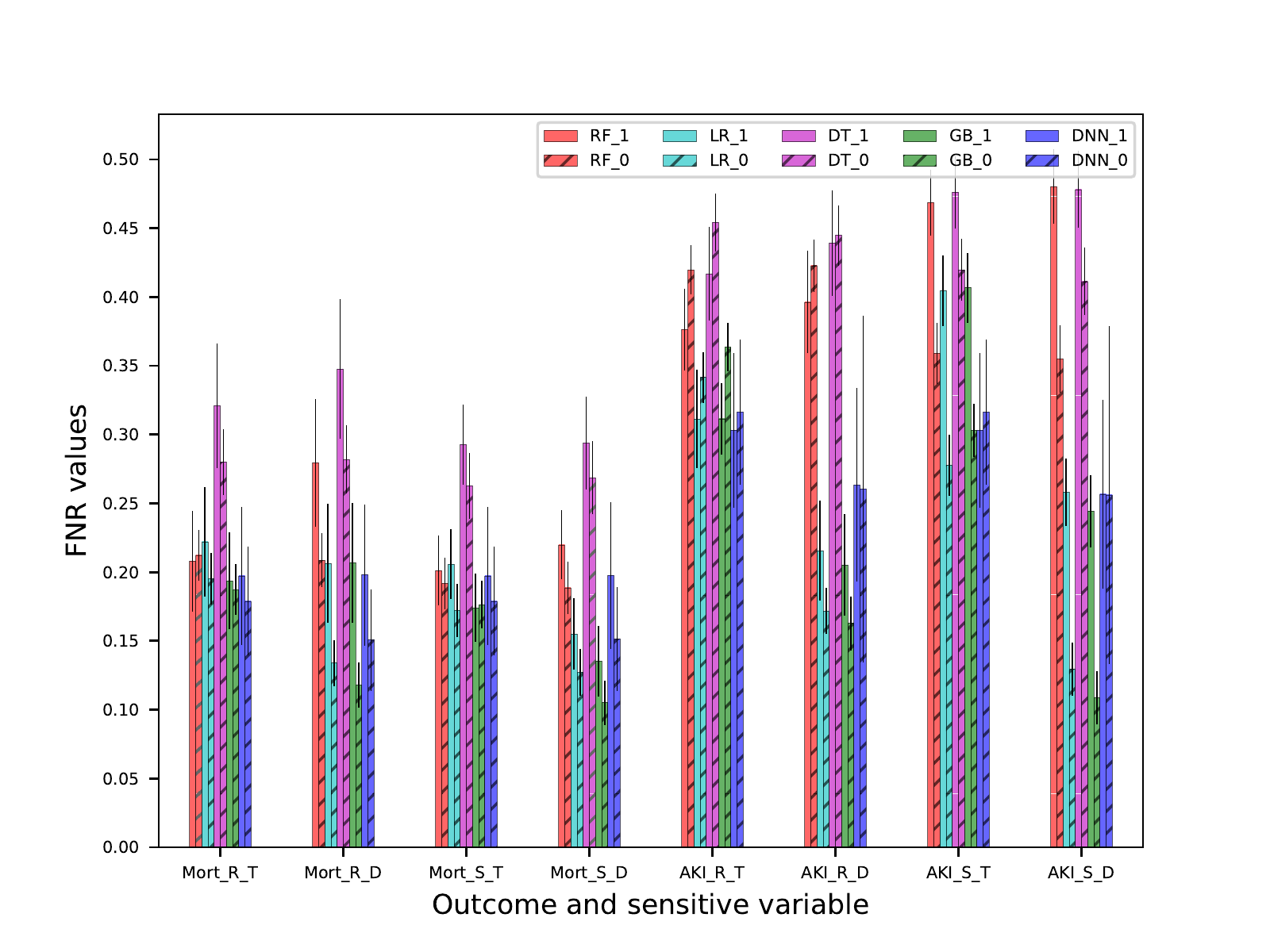}
    \end{subfigure}
    \vskip\baselineskip
    \vspace{-0.5cm}
    \begin{subfigure}[b]{0.9\textwidth}  
        \centering 
        \includegraphics[width=20cm, height =11cm]{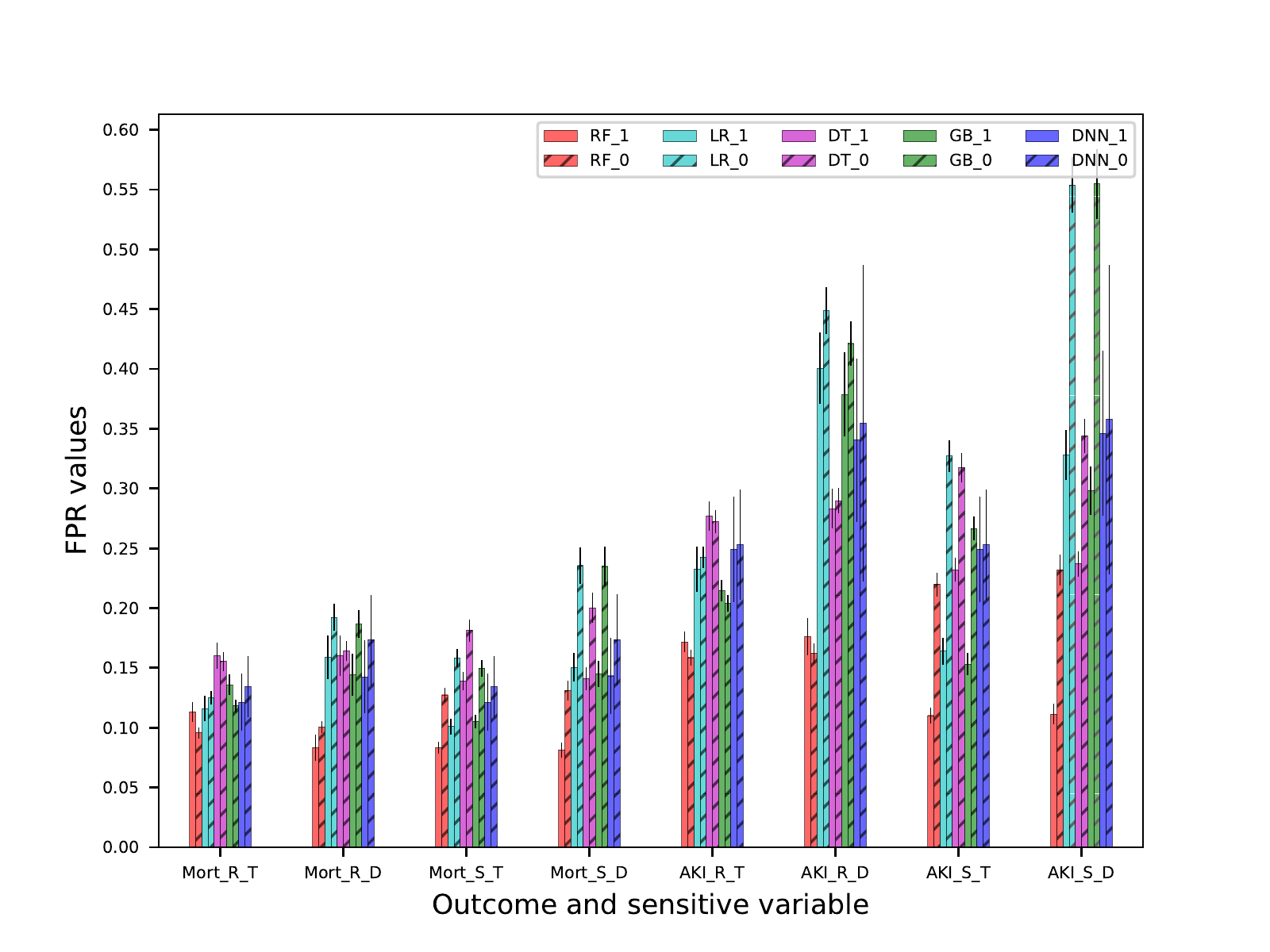}
    \end{subfigure}
    \caption{Comparison of performance measure $\in$ {FNR, FPR} values for the different classifiers  marked by color across sensitive characteristic race (R) or sex (S) subgroups. Some classifiers were trained separately (D) on subgroups of a particular value of sensitive characteristic and others on combined subgroup dat (T). }
    \label{figapp: per_on_sep_groups_FPR_FNR}
\end{figure*}
\begin{figure*}[h!]
    \centering
    \begin{subfigure}[b]{0.9\textwidth}
        \centering
        \includegraphics[width=20cm, height =11cm]{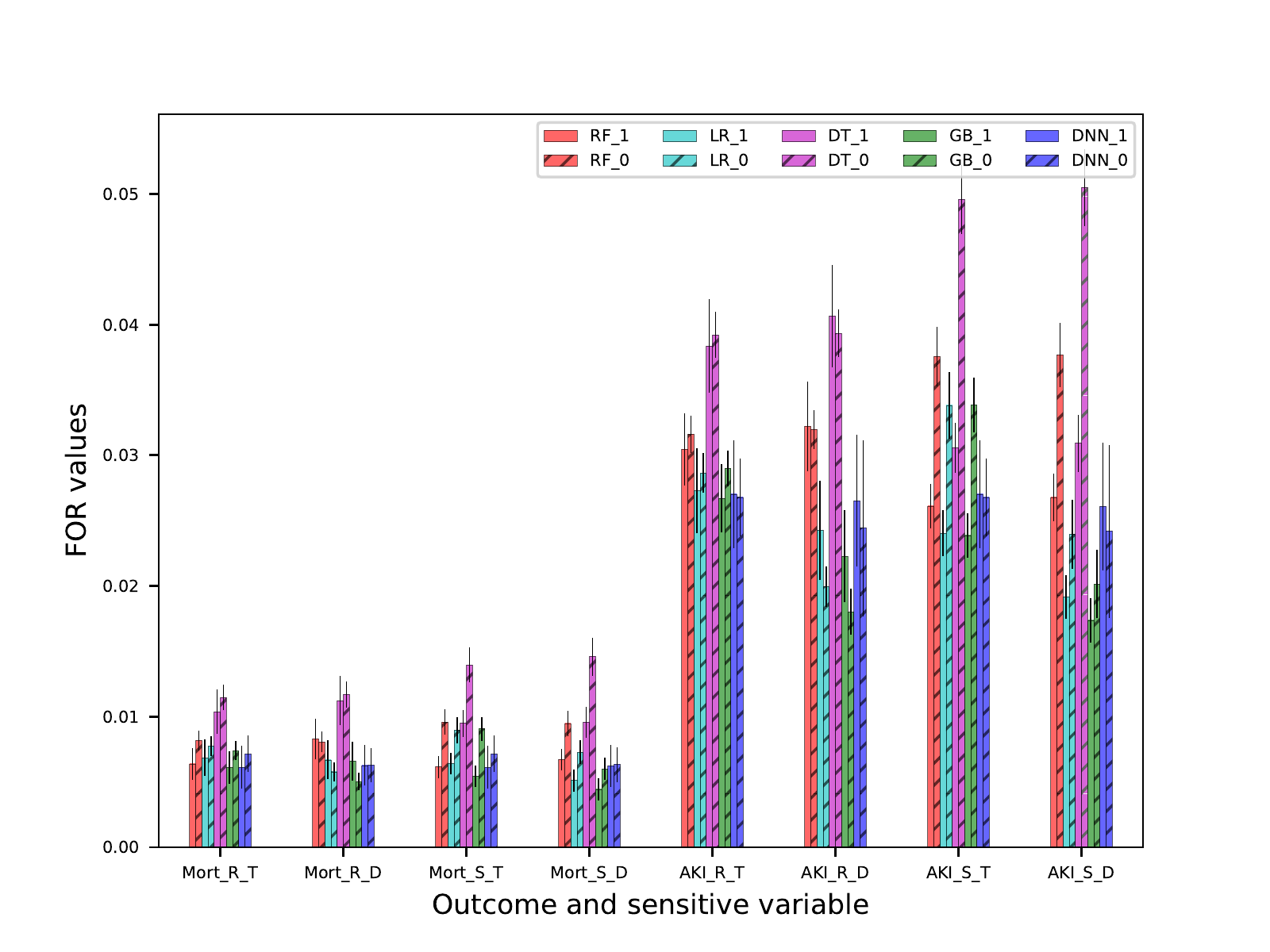}
    \end{subfigure}
    \vskip\baselineskip
    \vspace{-0.5cm}
    \begin{subfigure}[b]{0.9\textwidth}  
        \centering 
        \includegraphics[width=20cm, height =11cm]{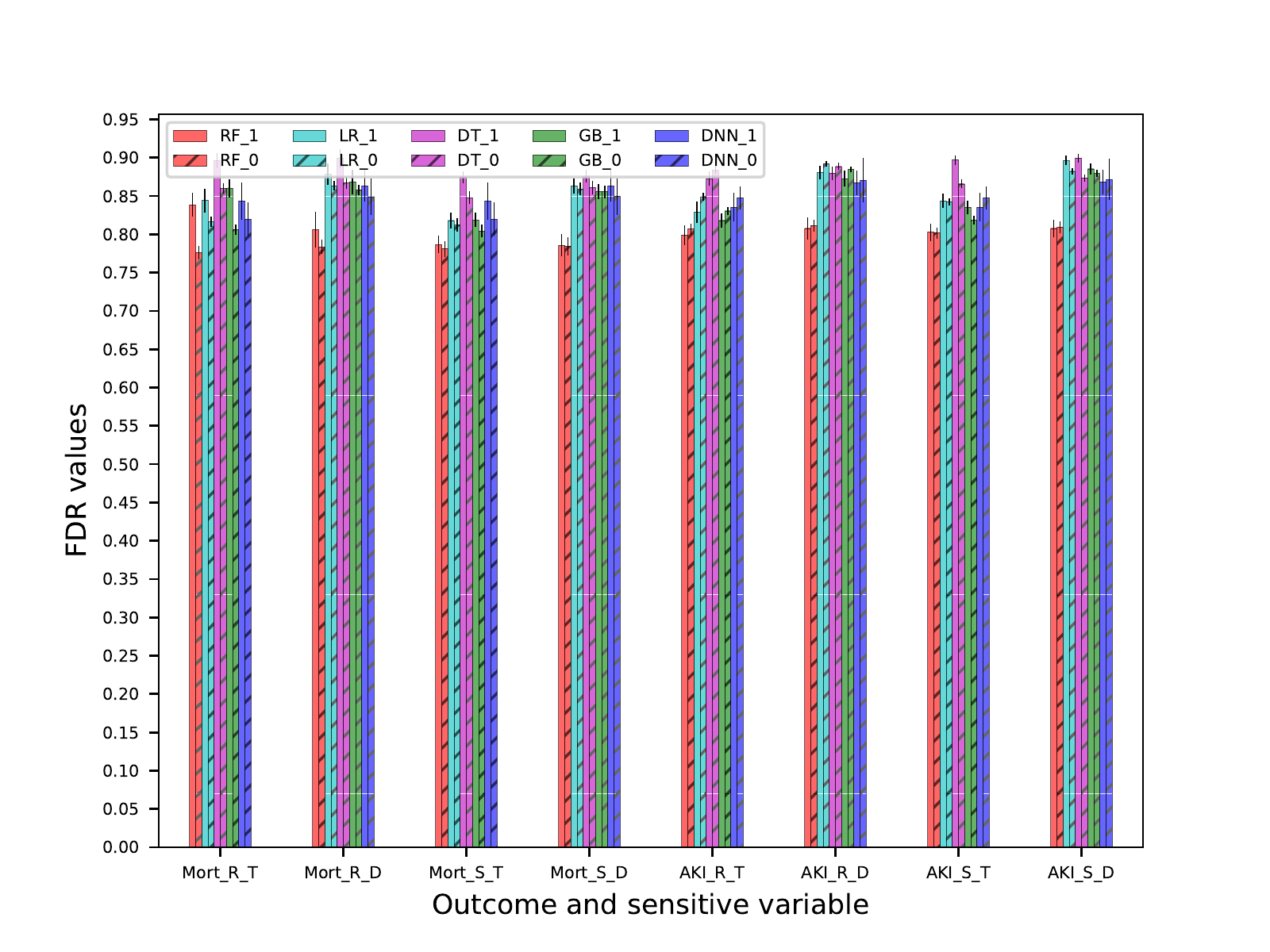}
    \end{subfigure}
    \caption{Comparison of performance measure $\in$ {FOR, FDR} values for the different classifiers  marked by color across sensitive characteristic race (R) or sex (S) subgroups. Some classifiers were trained separately (D) on subgroups of a particular value of sensitive characteristic and others on combined subgroup data (T). }
    \label{figapp: per_on_sep_groups_FOR_FDR}
\end{figure*}


\begin{figure*}[h!]
    \centering
    \begin{subfigure}[b]{0.9\textwidth}
        \centering
        \includegraphics[width=20cm, height =11cm]{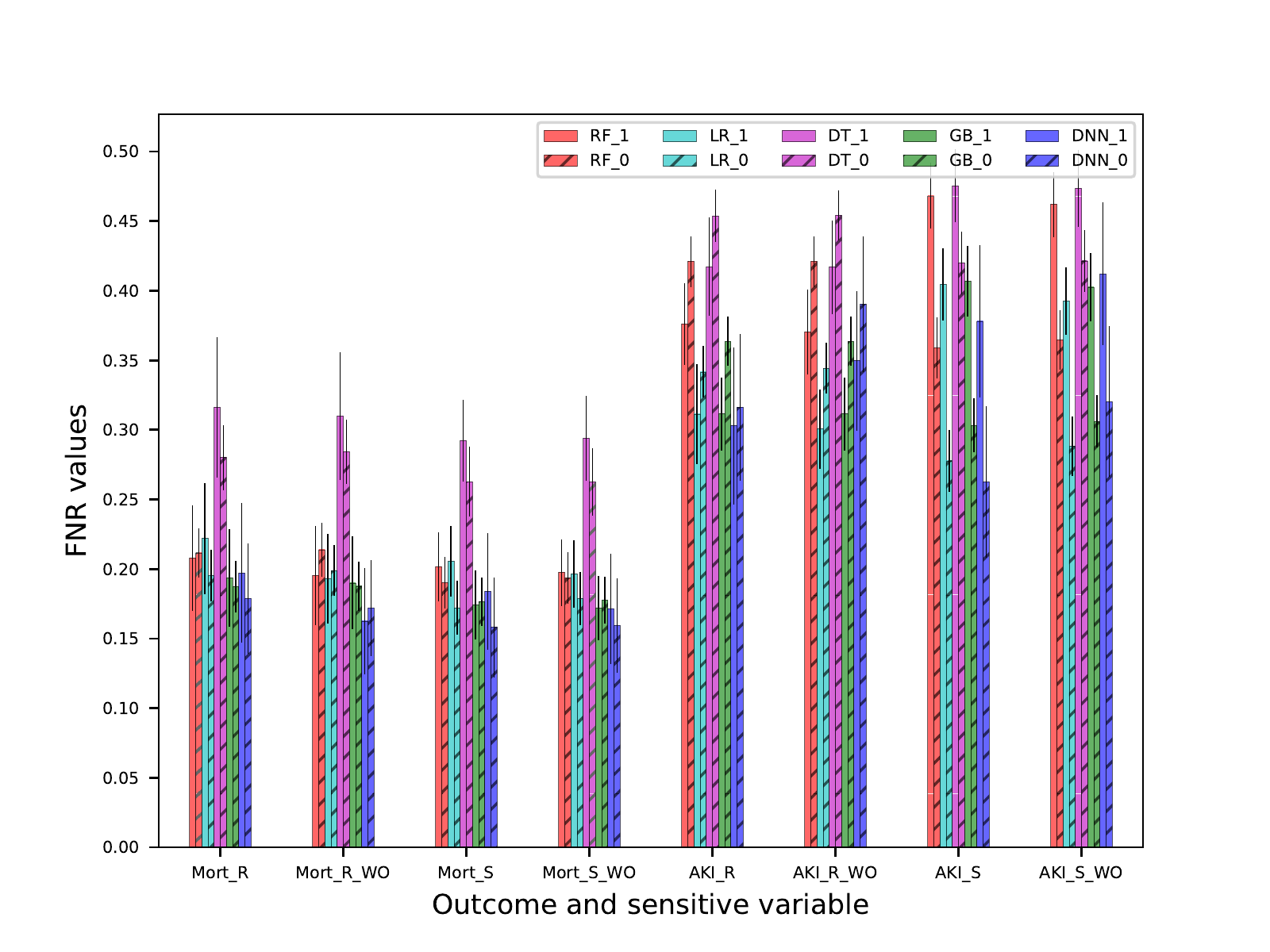}
    \end{subfigure}
    \vskip\baselineskip
    \vspace{-0.5cm}    \begin{subfigure}[b]{0.9\textwidth}  
        \centering 
        \includegraphics[width=20cm, height =11cm]{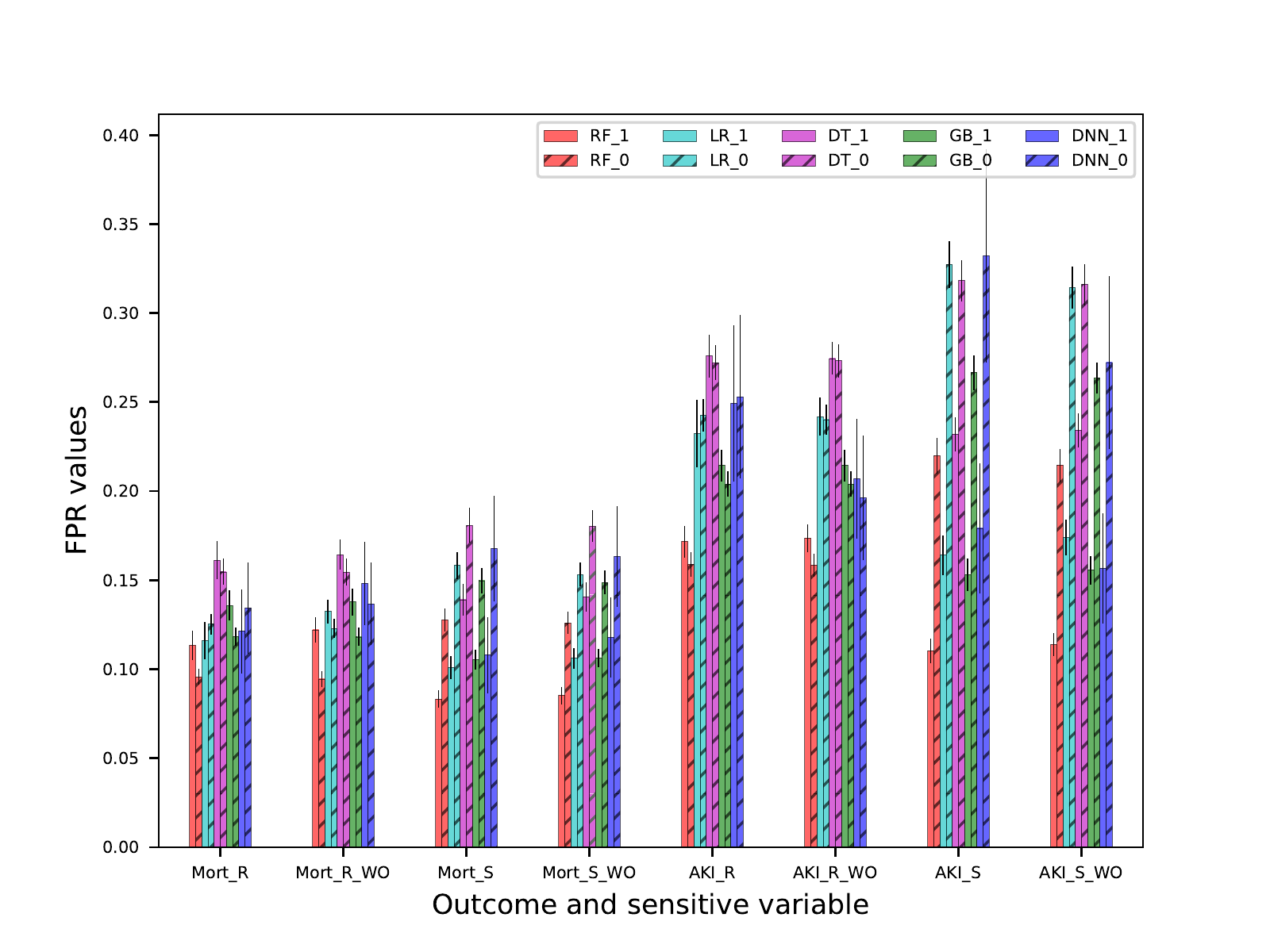}
    \end{subfigure}
    \caption{Comparison of performance measures $\in$\{FNR, FPR\} for the cases when the sensitive variable was used or not (WO) while training across race (R) and sex (S). As can be seen, FNR differences decreased but the FPR differences increased even though the magnitude is very small. Overall, removing the sensitive characteristic doesn't decrease the bias.}
    \label{figapp: basic_WO_models_diff_perf_FPR_FNR}
\end{figure*}

\begin{figure*}[h!]
    \centering
    \begin{subfigure}[b]{0.9\textwidth}
        \centering
        \includegraphics[width=20cm, height =11cm]{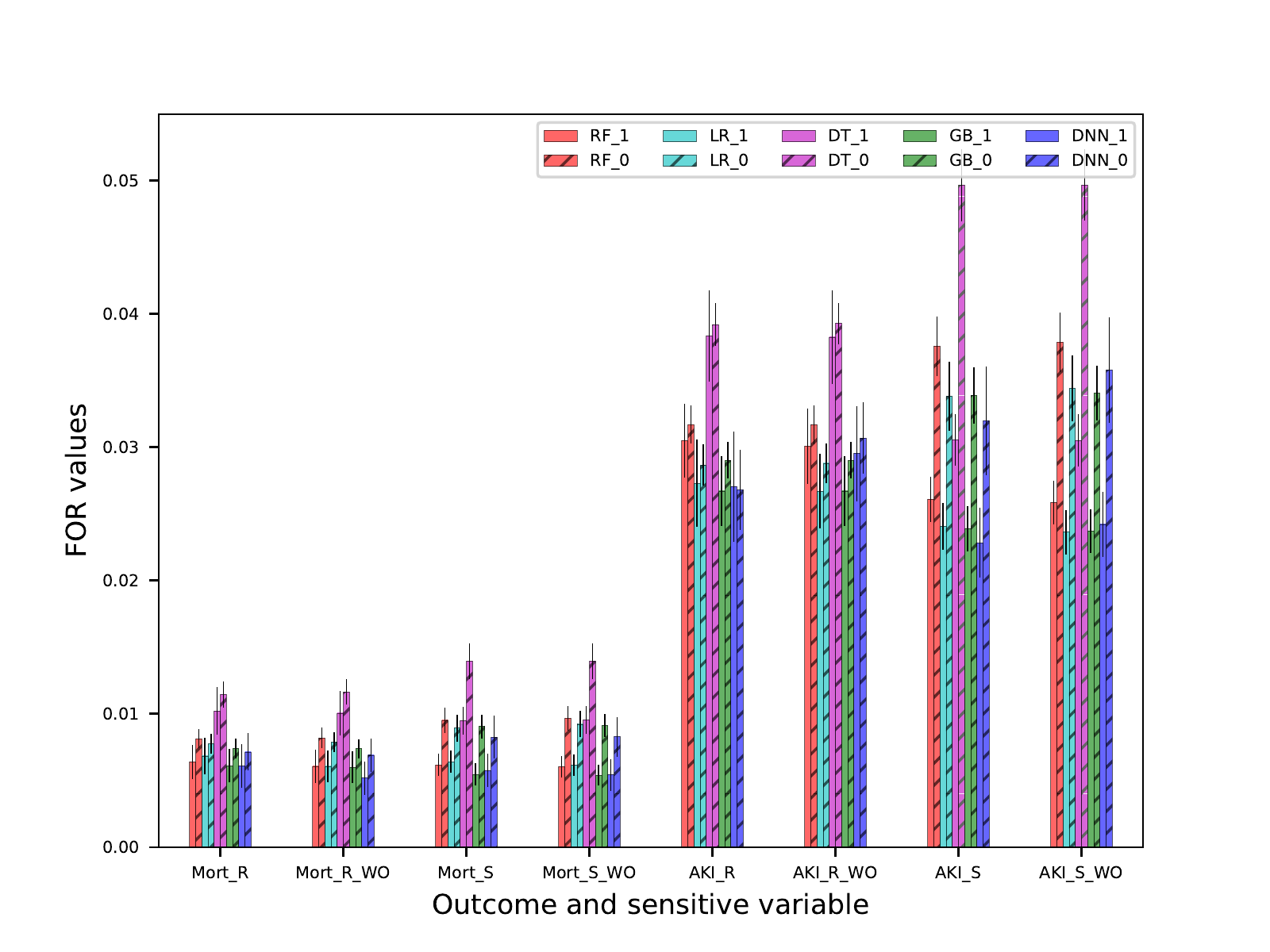}
    \end{subfigure}
    \vskip\baselineskip
    \vspace{-0.5cm}    \begin{subfigure}[b]{0.9\textwidth}  
        \centering 
        \includegraphics[width=20cm, height =11cm]{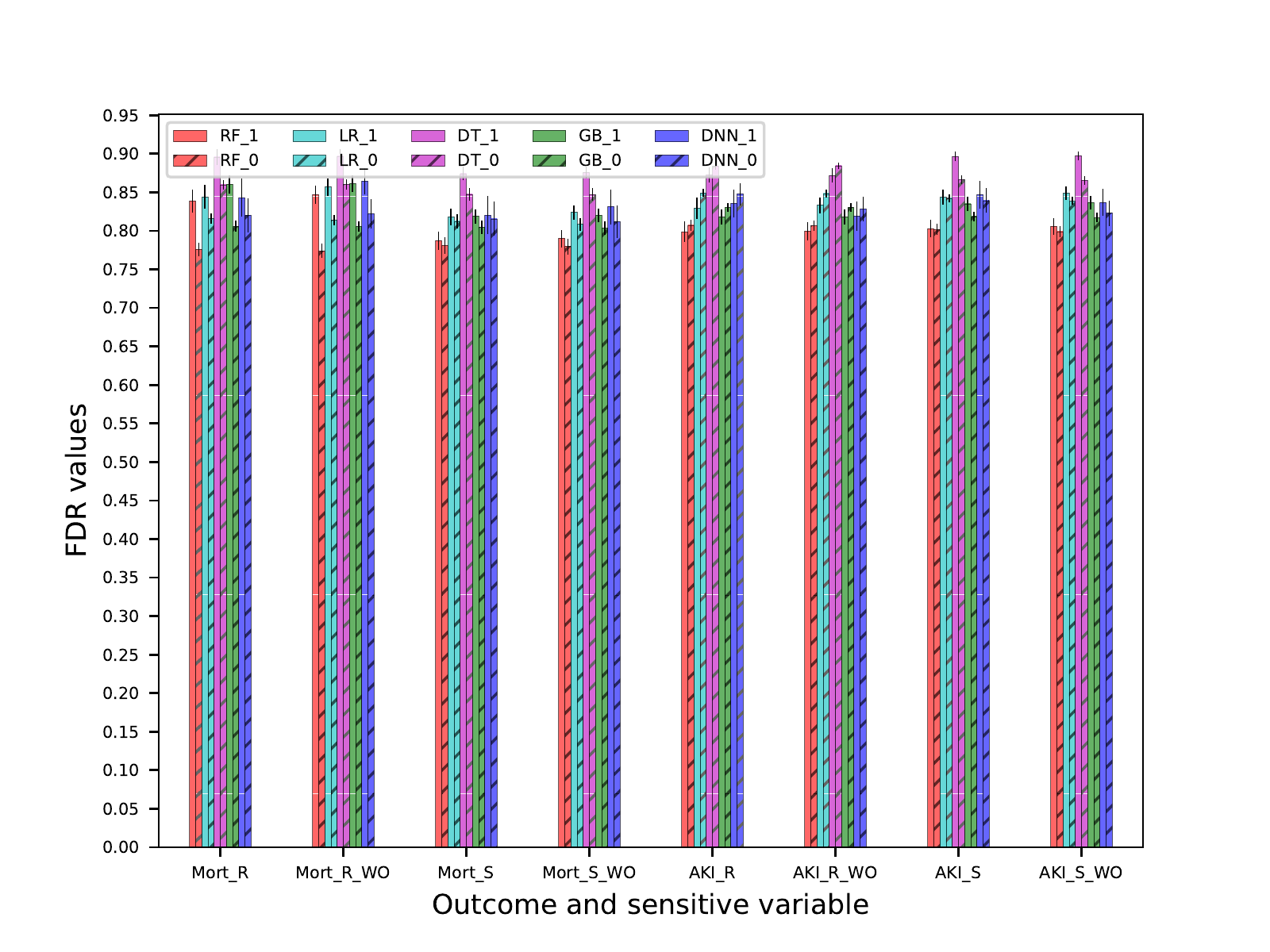}
    \end{subfigure}
    \caption{Comparison of performance measures $\in$\{FOR, FDR\} for the cases when the sensitive variable was used or not (WO) while training across race (R) and sex (S). As can be seen, there is almost no change in the differences implying that removing the sensitive characteristic doesn't decrease the bias.}
    \label{figapp: basic_WO_models_diff_perf_FOR_FDR}
\end{figure*}


\begin{figure*}[h!]
    \centering
    \begin{subfigure}[b]{0.9\textwidth}
        \centering
        \includegraphics[width=20cm, height =11cm]{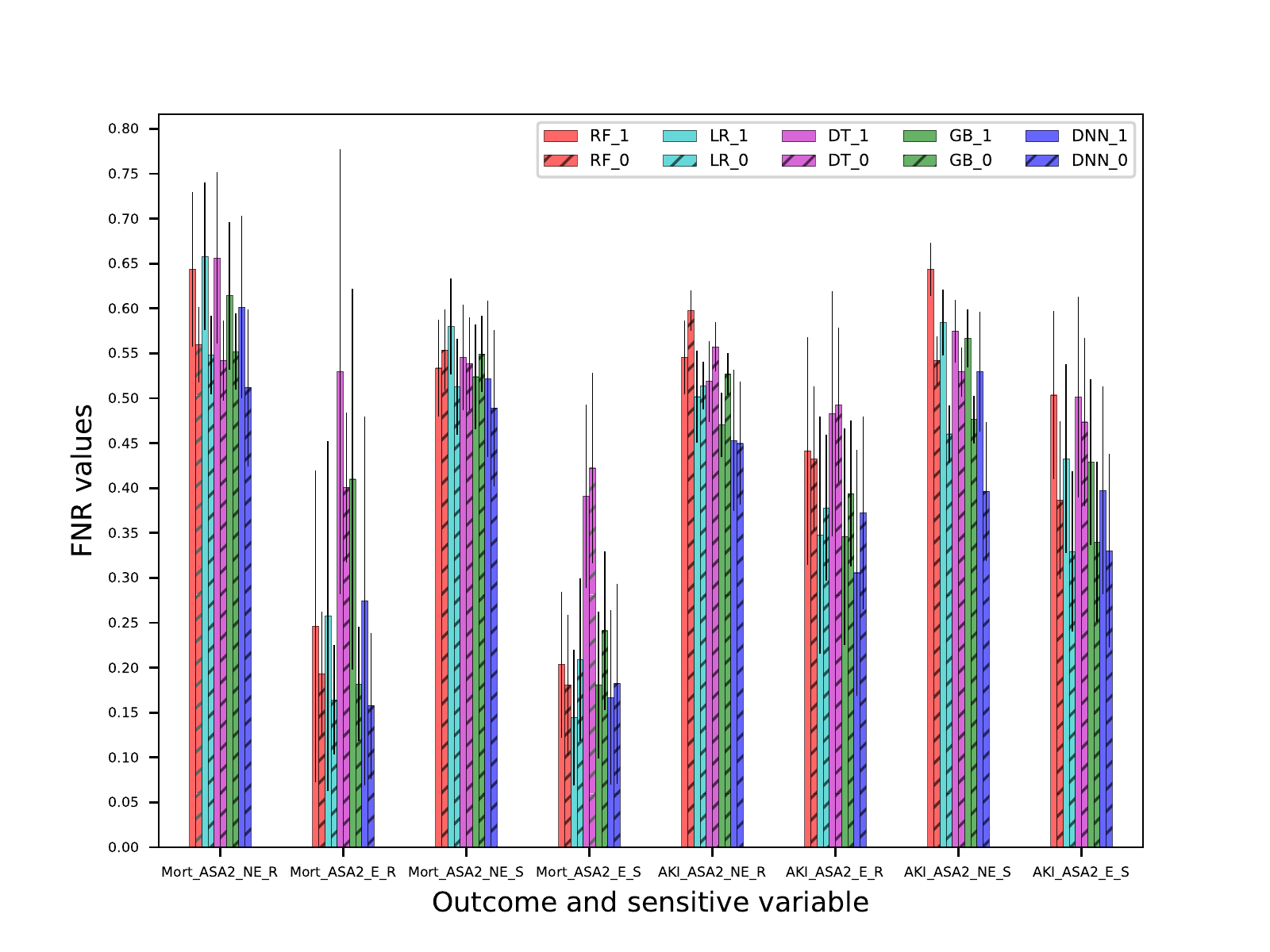}
    \end{subfigure}
    \hfill
    \begin{subfigure}[b]{0.9\textwidth}  
        \centering 
        \includegraphics[width=20cm, height =11cm]{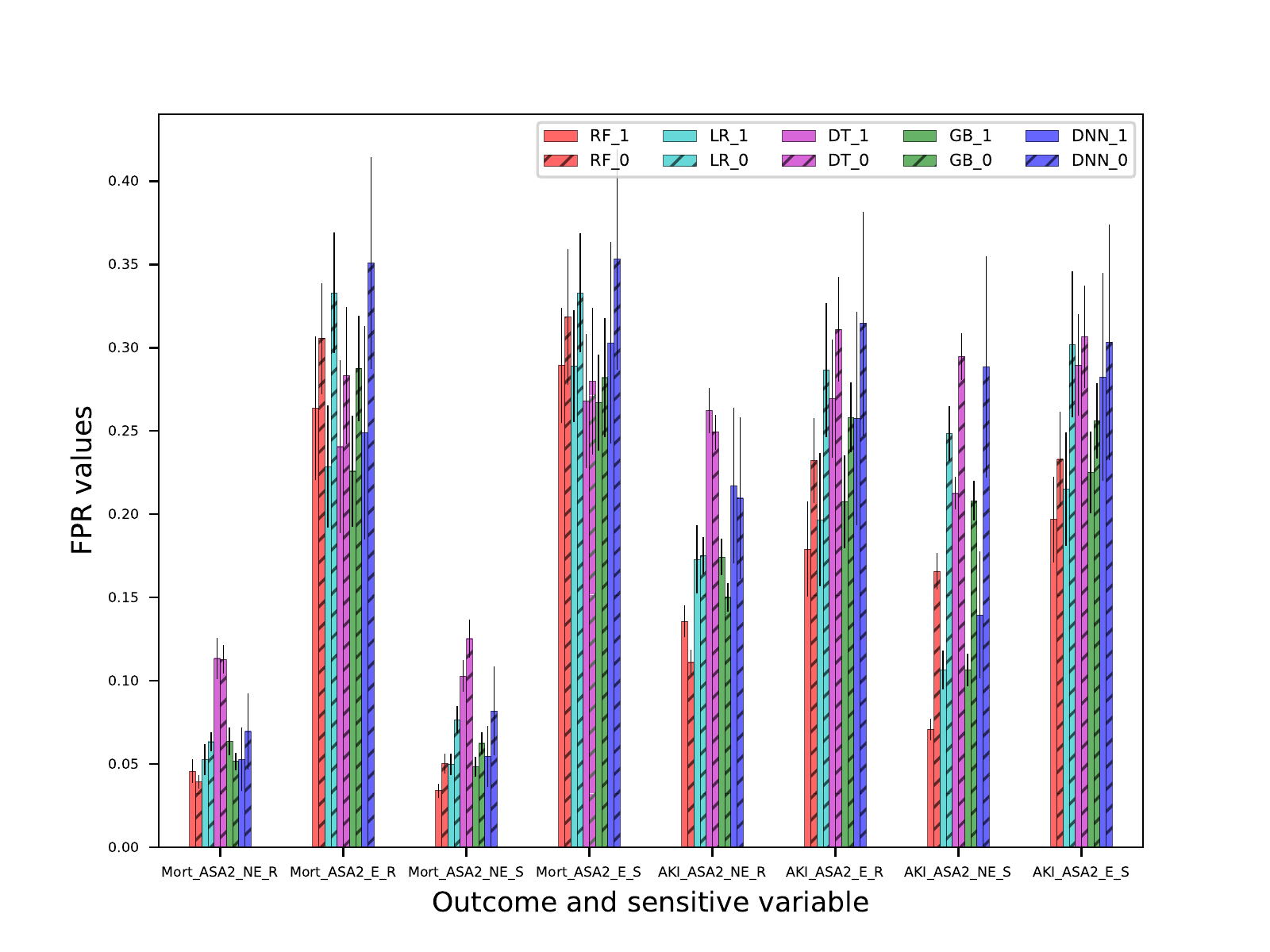}
    \end{subfigure}
    \caption{Comparison of performance measure $\in$ \{FNR, FPR\} values on the group ASA2\_NE and ASA2\_E (patients having systematic disease) across the values taken by sensitive characteristics race (R) and sex (S). E and NE denotes the group categorized in ASA emergency case or not.}
    \label{figapp: strat_ASA2_FNFR_FPR}
\end{figure*}

\begin{figure*}[h!]
    \centering
    \begin{subfigure}[b]{0.9\textwidth}
        \centering
        \includegraphics[width=20cm, height =11cm]{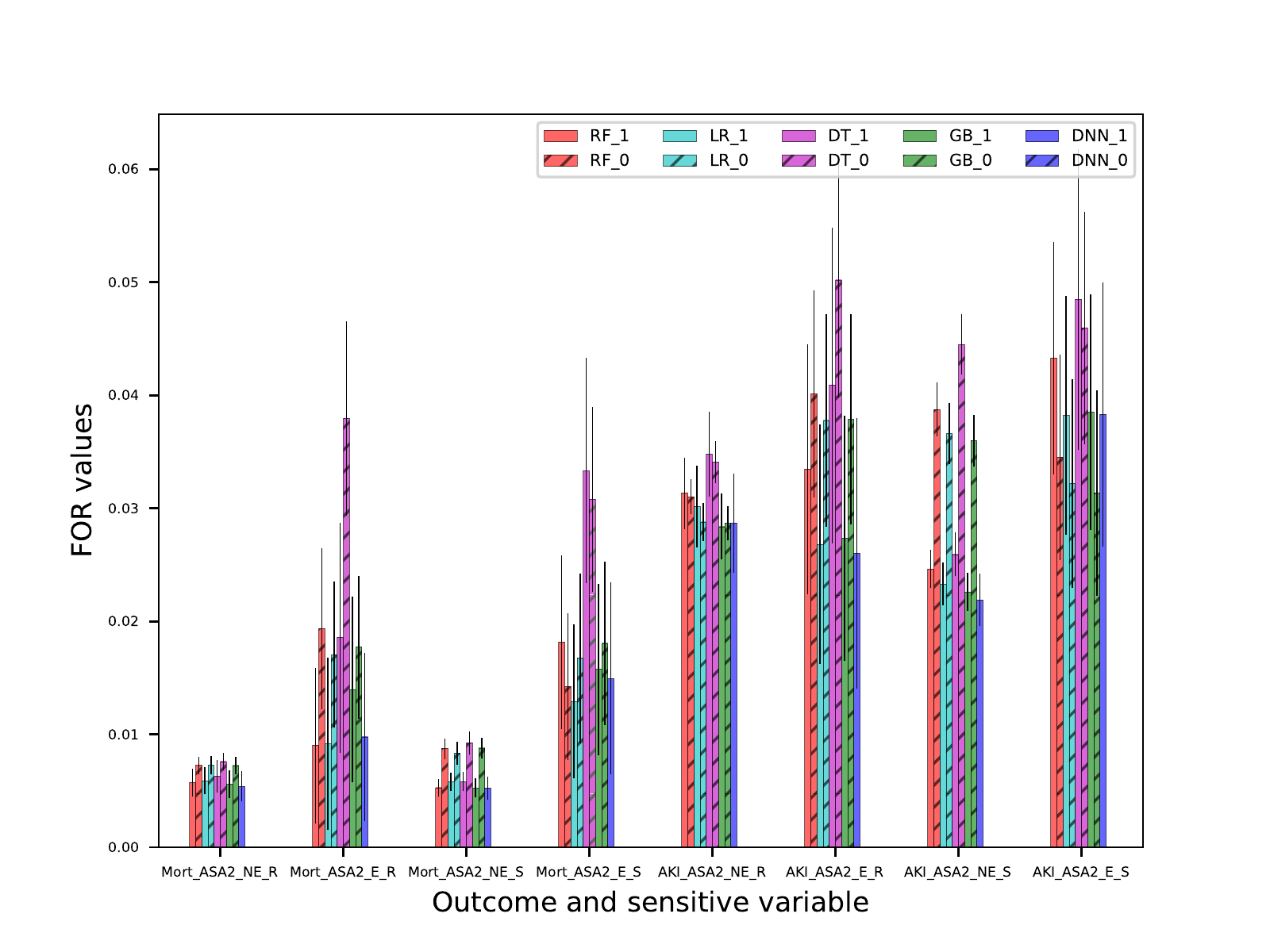}
    \end{subfigure}
    \hfill
    \begin{subfigure}[b]{0.9\textwidth}  
        \centering 
        \includegraphics[width=20cm, height =11cm]{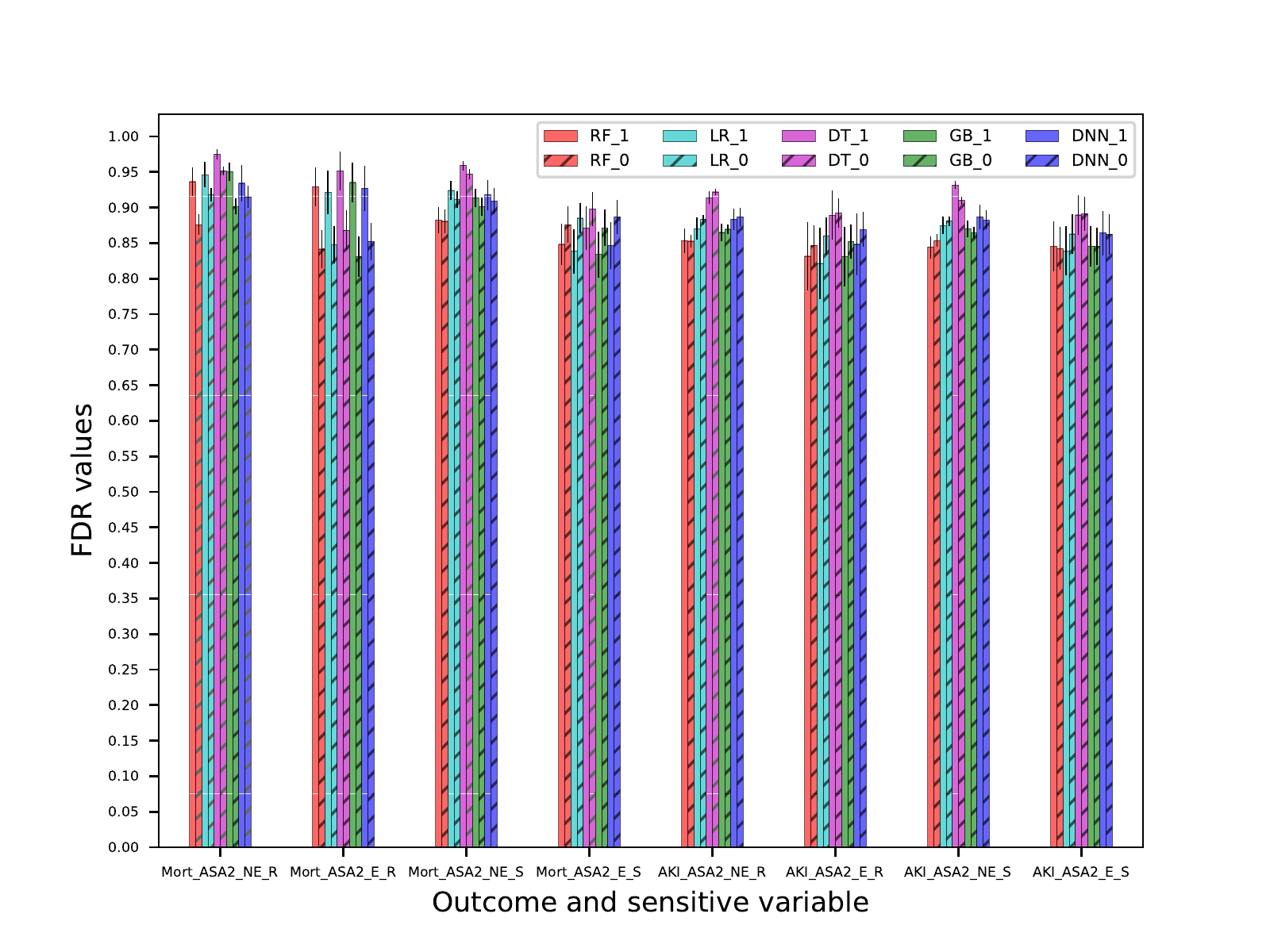}
    \end{subfigure}
    \caption{Comparison of performance measure $\in$ \{FOR, FDR\} values on the group ASA2\_NE and ASA2\_E (patients having systematic disease) across the values taken by sensitive characteristics race (R) and sex (S). E and NE denotes the group categorized in ASA emergency case or not.}
    \label{figapp: strat_ASA2_FOR_FDR}
\end{figure*}

\begin{figure*}[h!]
    \centering
    \begin{subfigure}[b]{0.9\textwidth}
        \centering
        \includegraphics[width=20cm, height =11cm]{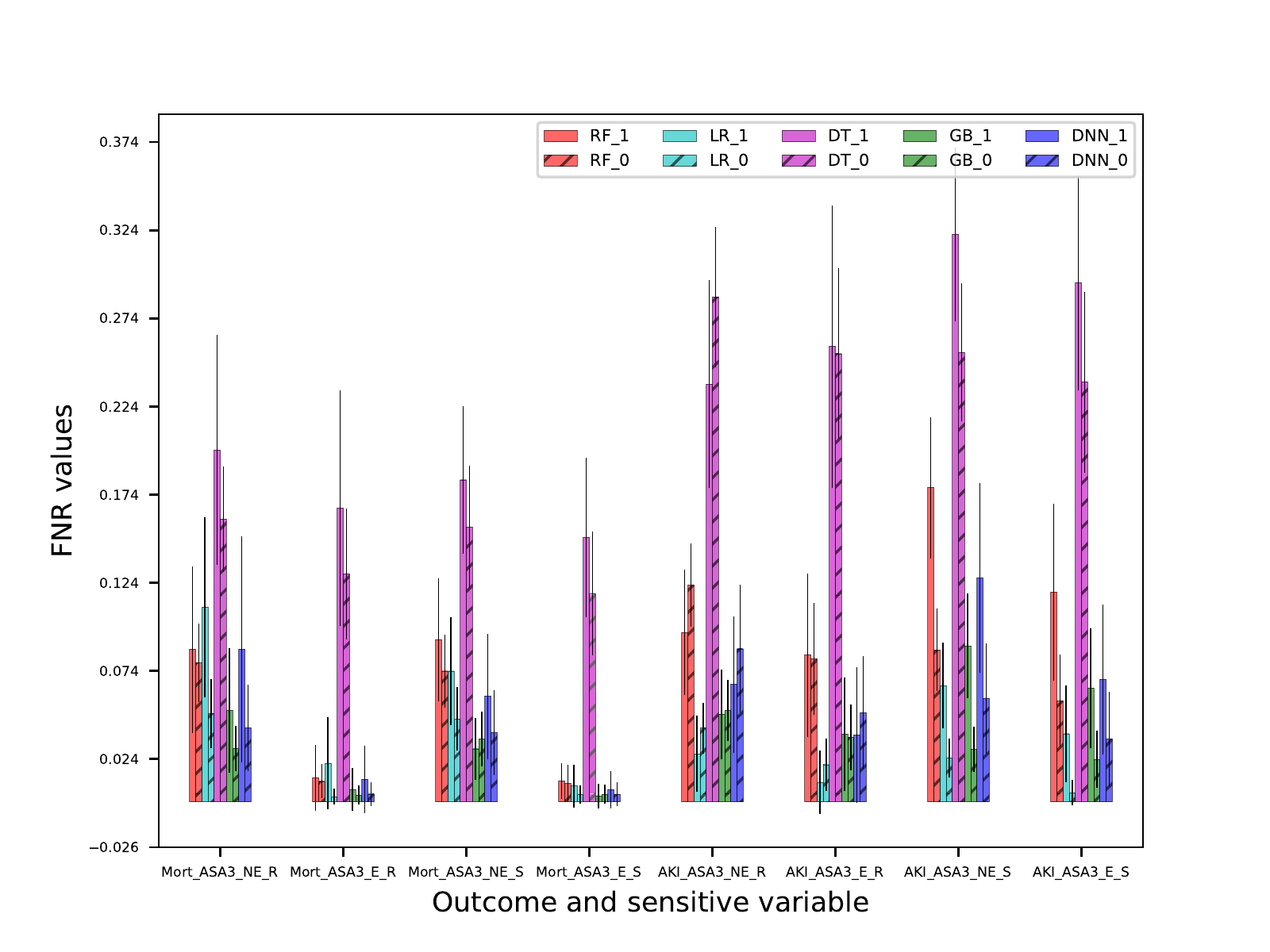}
    \end{subfigure}
    \hfill
    \begin{subfigure}[b]{0.9\textwidth}  
        \centering 
        \includegraphics[width=20cm, height =11cm]{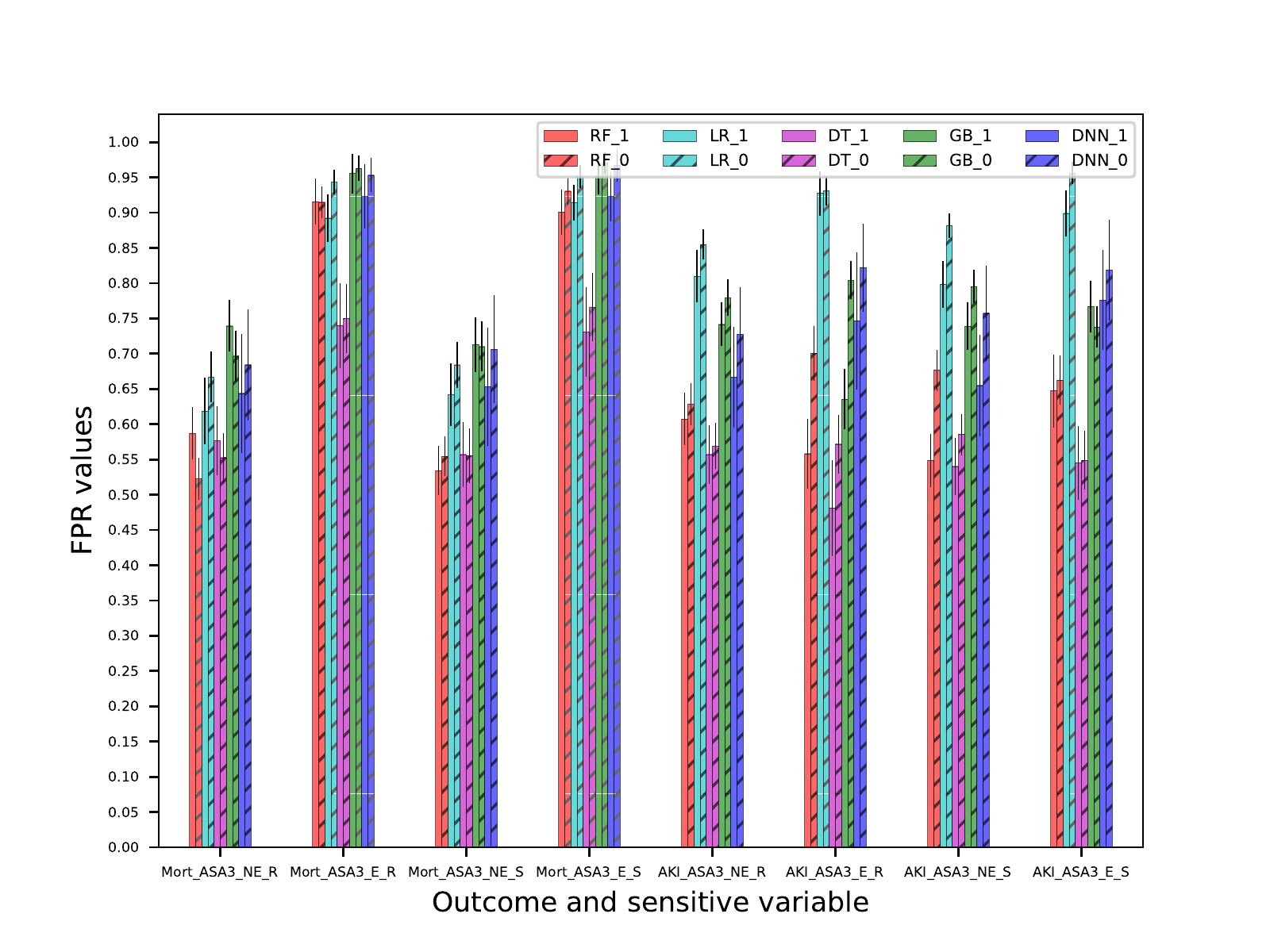}
    \end{subfigure}
    \caption{Comparison of performance measures $\in$ \{FNR, FPR\} values on the group ASA3\_NE and ASA3\_E (patients having constant threat to life) across the values taken by sensitive characteristics race (R) and sex (S). E and NE denotes the group categorized in ASA emergency case or not.}
    \label{figapp: strat_ASA3_FNR_FPR}
\end{figure*}

\begin{figure*}[h!]
    \centering
    \begin{subfigure}[b]{0.9\textwidth}
        \centering
        \includegraphics[width=20cm, height =11cm]{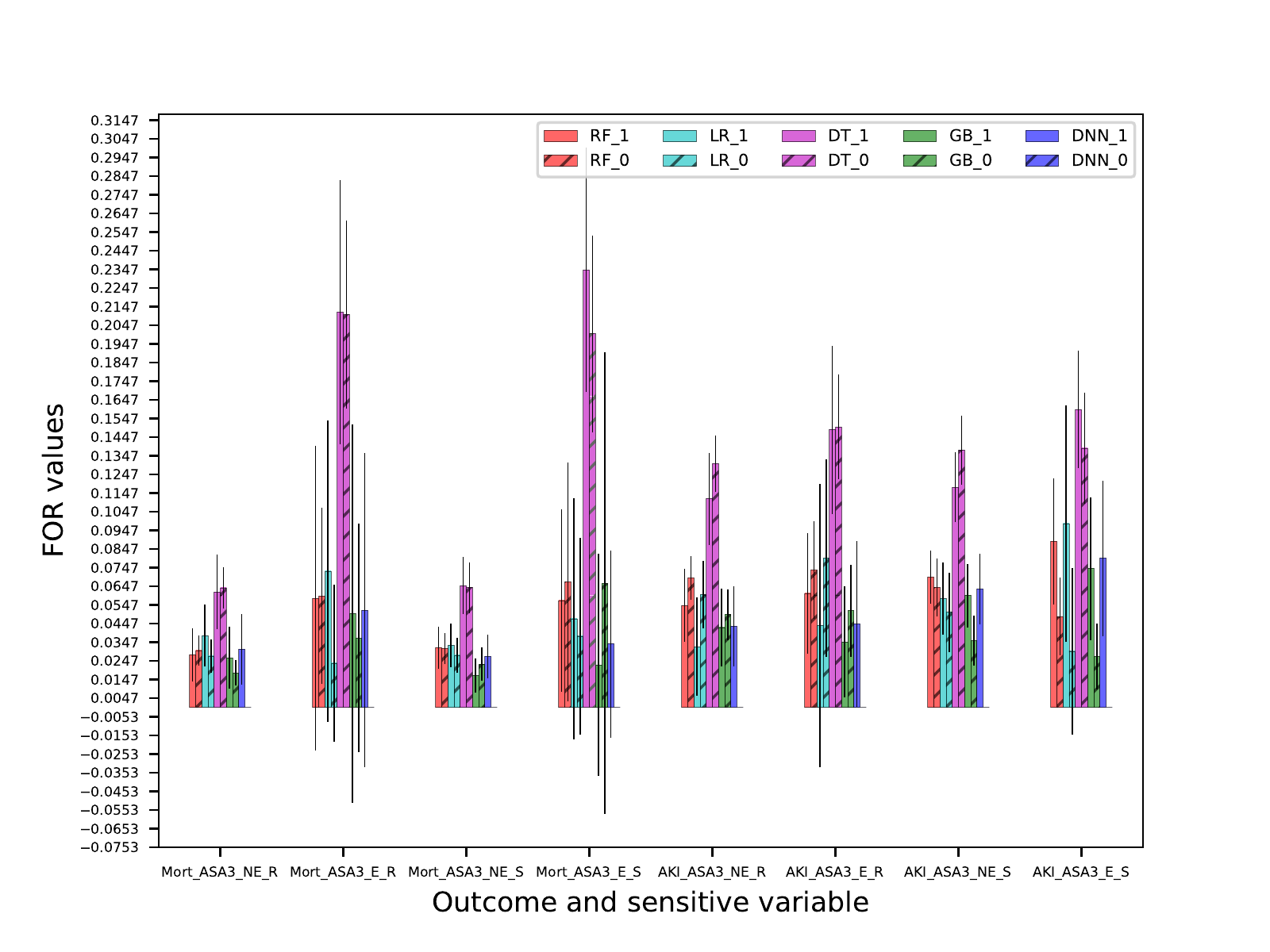}
    \end{subfigure}
    \hfill
    \begin{subfigure}[b]{0.9\textwidth}  
        \centering 
        \includegraphics[width=20cm, height =11cm]{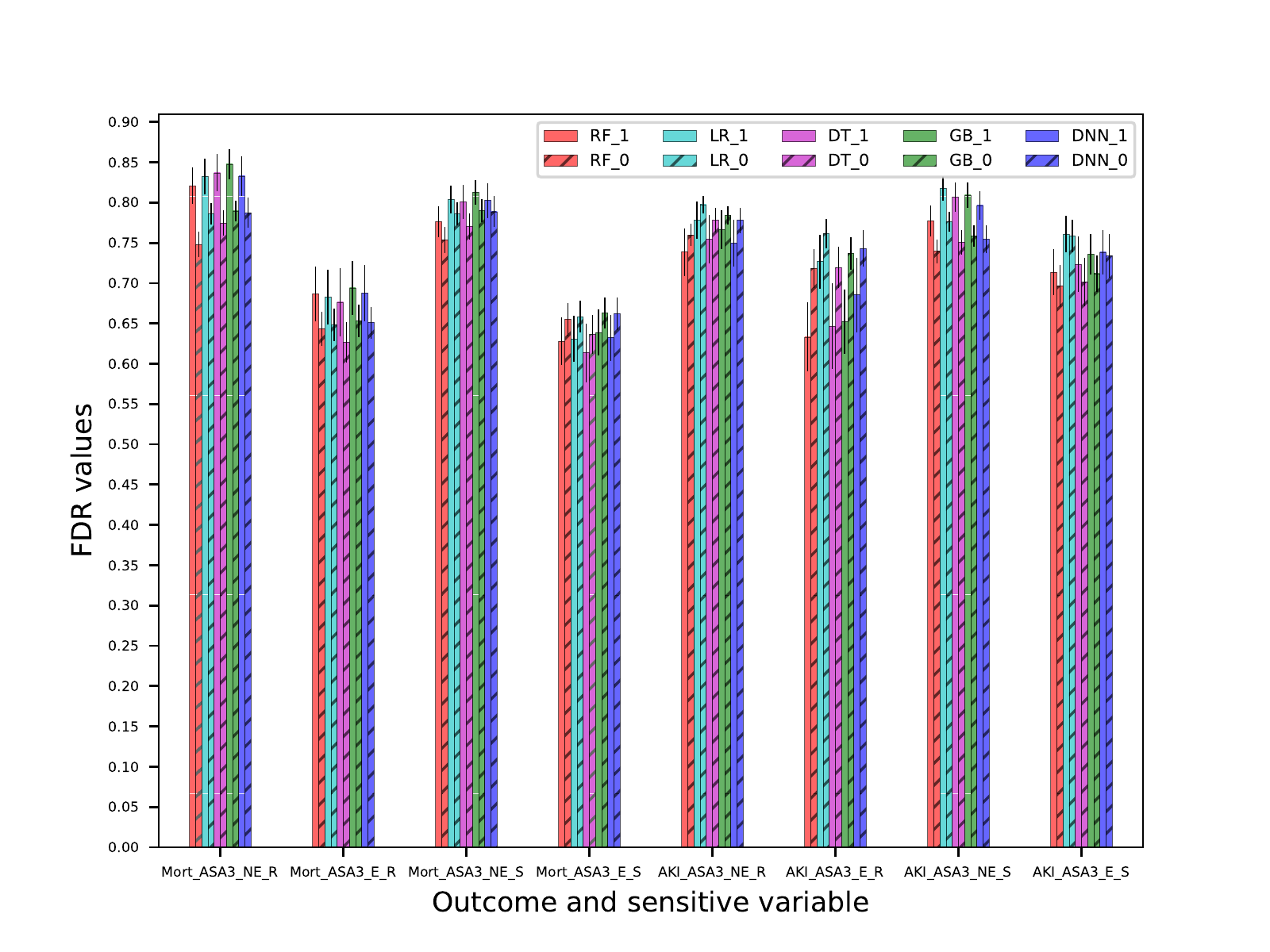}
    \end{subfigure}
    \caption{Comparison of performance measures $\in$ \{FOR, FDR\} values on the group ASA3\_NE and ASA3\_E (patients having constant threat to life) across the values taken by sensitive characteristics race (R) and sex (S). E and NE denotes the group categorized in ASA emergency case or not.}
    \label{figapp: strat_ASA3_FOR_FDR}
\end{figure*}

\begin{figure*}[h!]
    \centering
    \begin{subfigure}[b]{0.9\textwidth}
        \centering
        \includegraphics[width=20cm, height =11cm]{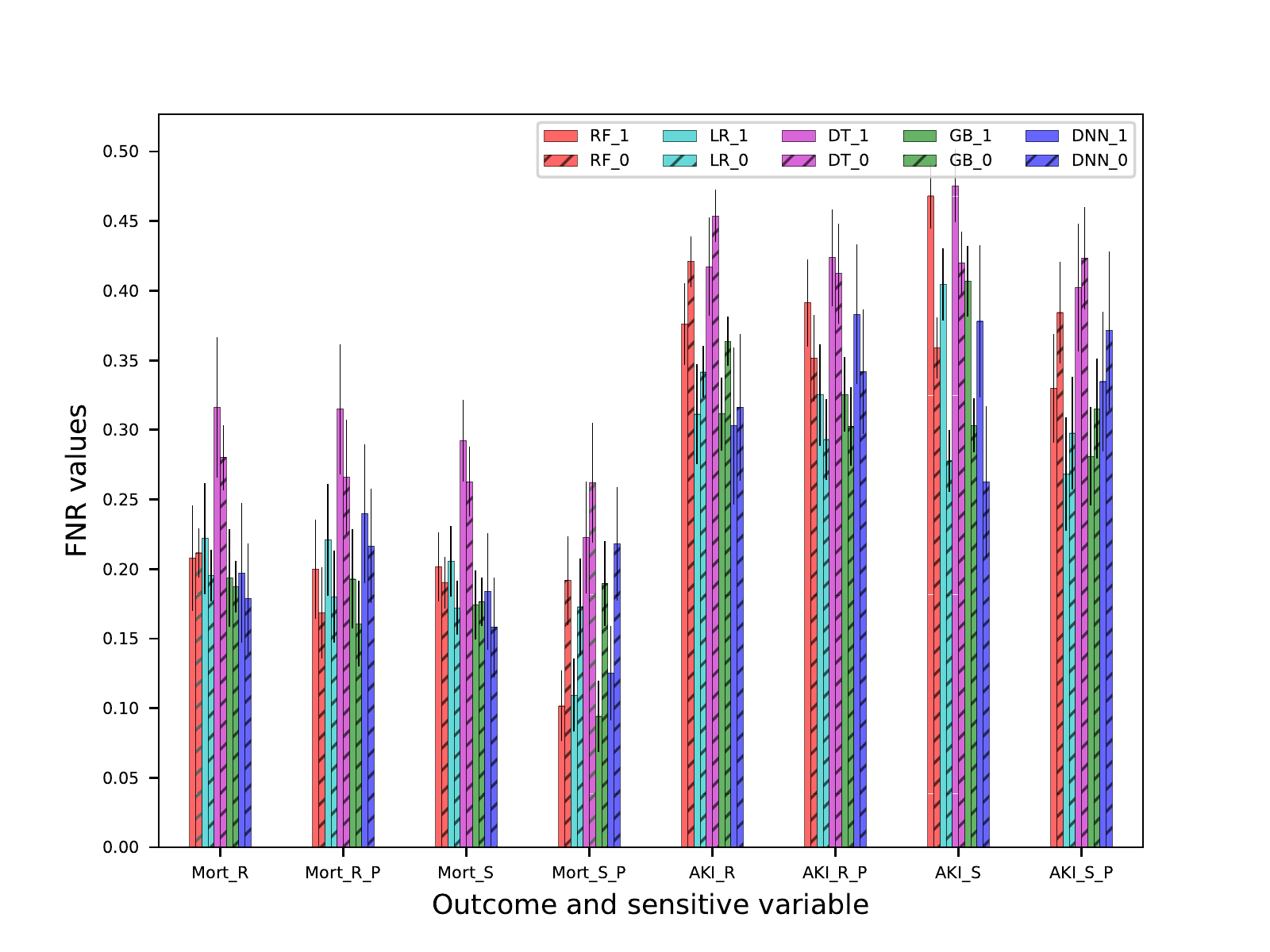}
    \end{subfigure}
    \hfill
    \begin{subfigure}[b]{0.9\textwidth}  
        \centering 
        \includegraphics[width=20cm, height =11cm]{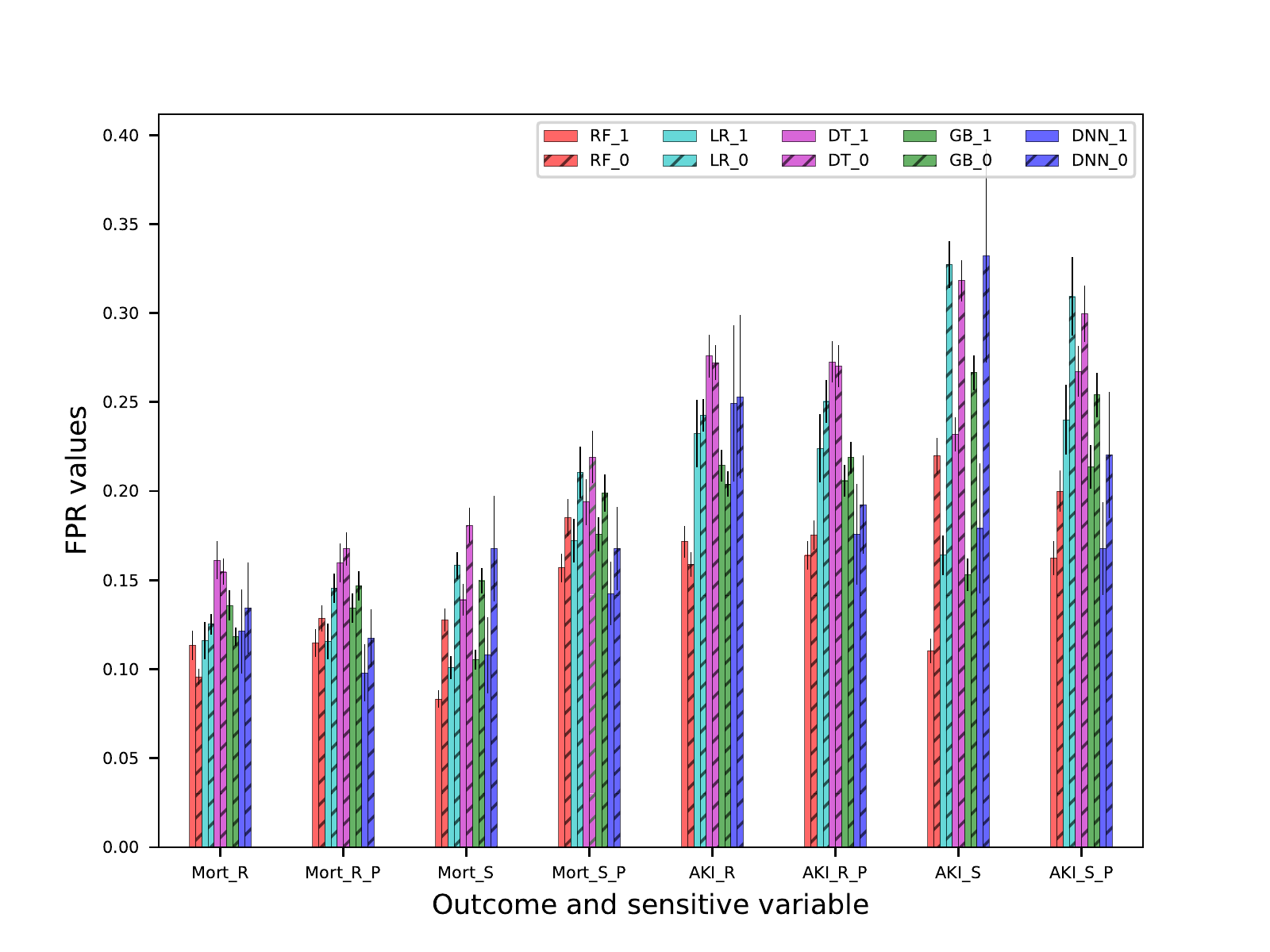}
    \end{subfigure}
    \caption{Comparison of performance measures values $\in$ \{FNR, FPR\} on complete test set and  propensity matched test set (P) across the subgroups of sensitive characteristics Race (R) and Sex (S).}
    \label{figapp: per_on_PSM_set_FNR_FPR}
\end{figure*}

\begin{figure*}[h!]
    \centering
    \begin{subfigure}[b]{0.9\textwidth}
        \centering
        \includegraphics[width=20cm, height =11cm]{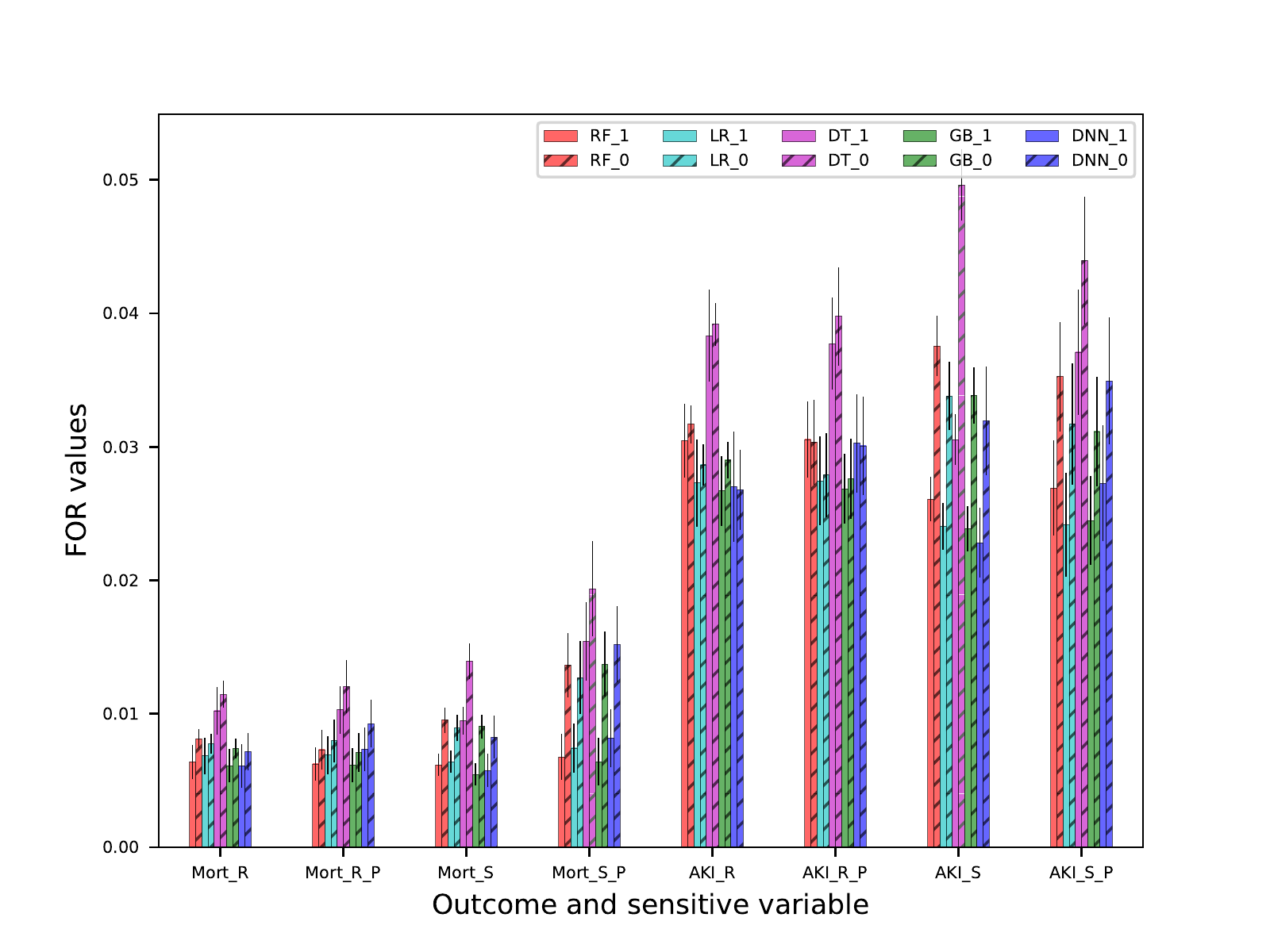}
    \end{subfigure}
    \hfill
    \begin{subfigure}[b]{0.9\textwidth}  
        \centering 
        \includegraphics[width=20cm, height =11cm]{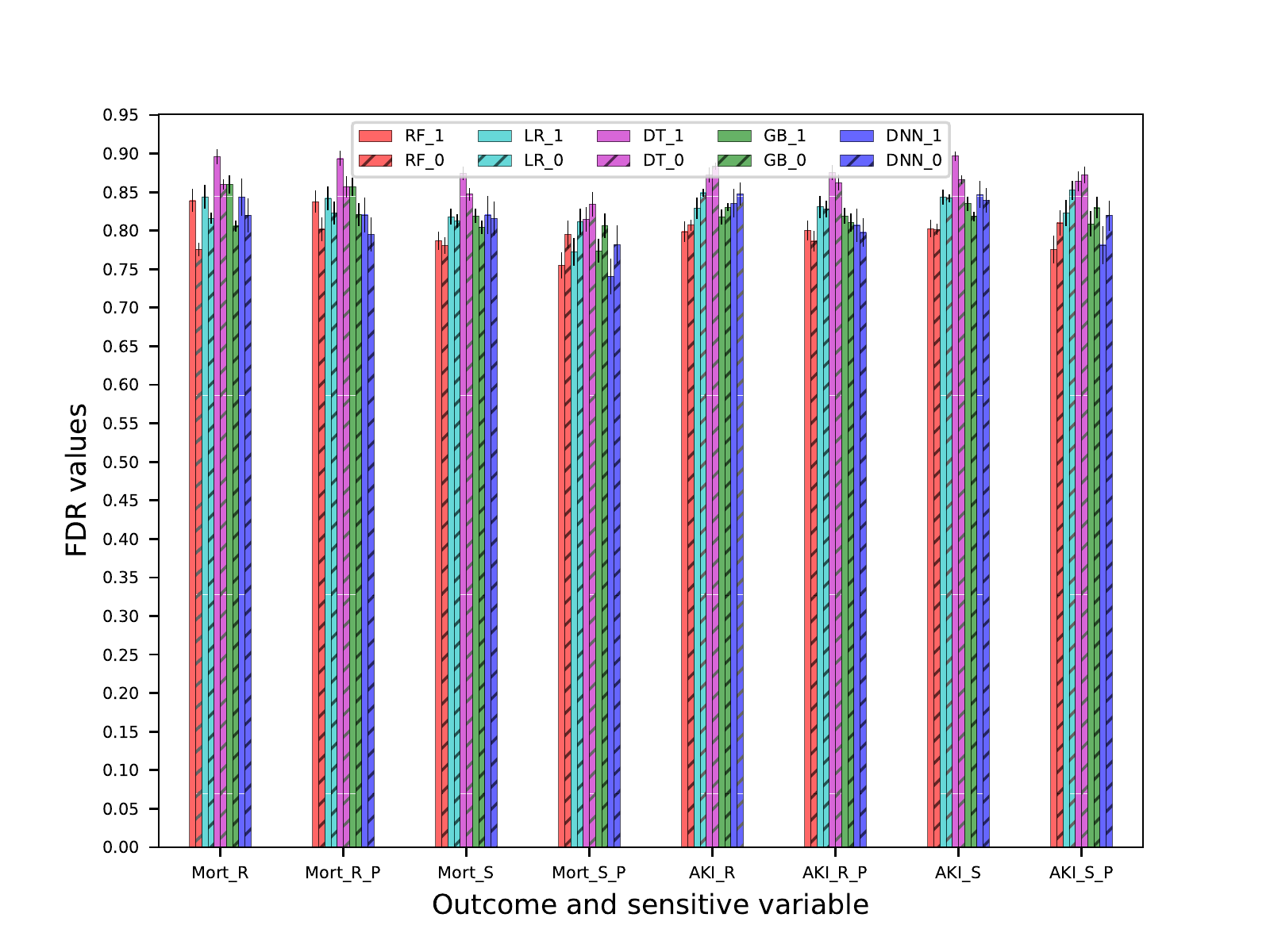}
    \end{subfigure}
    \caption{Comparison of performance measures values $\in$ \{FOR, FDR\} on complete test set and  propensity matched test set (P) across the subgroups of sensitive characteristics Race (R) and Sex (S).}
    \label{figapp: per_on_PSM_set_FOR_FDR}
\end{figure*}
\end{document}